\documentclass{article}
\usepackage{times}
\usepackage{pict2e}

\usepackage{graphicx} % more modern
\usepackage{subfigure}
\usepackage{bm}
\usepackage{url}
\usepackage{amssymb}
\usepackage{soul,color}
\usepackage{graphicx,float,wrapfig}
\usepackage{amsmath,amsfonts,amsthm,amssymb}
\usepackage{xr}
\usepackage[english]{babel}
\usepackage{multirow}
\usepackage{slashbox}
\usepackage[table]{xcolor}

\usepackage{natbib}

\usepackage{algorithm}
\usepackage{algorithmic}

\usepackage{hyperref}

\newcommand{\xv}{\bm{x}}
\newcommand{\Xv}{\bm{X}}

\newcommand{\yv}{\bm{y}}
\newcommand{\Yv}{\bm{Y}}

\newcommand{\zv}{\bm{z}}
\newcommand{\Zv}{\bm{Z}}

\newcommand{\sv}{\bm{s}}
\newcommand{\Sv}{\bm{S}}

\newcommand{\hv}{\bm{h}}
\newcommand{\Hv}{\bm{H}}

\newcommand{\wv}{\bm{w}}

\newcommand{\prob}{\mathcal{P}}

\newcommand{\etav}{\bm \eta}
\newcommand{\thetav}{\bm{\theta}}
\newcommand{\Thetav}{\bm{\Theta}}
\newcommand{\phiv}{\bm{\phi}}
\newcommand{\Phiv}{\bm{\Phi}}
\newcommand{\muv}{\bm \mu}

\newcommand{\lambdav}{\bm \lambda}
\newcommand{\piv}{\bm \pi}
\newcommand{\barpiv}{\bm{\bar{\pi}}}

\newcommand{\Sigmav}{\bm \Sigma}
\newcommand{\Mv}{\bm{\mathcal{M}}}
\newcommand{\ep}{\mathbb{E}}
\newcommand{\KL}{\mathrm{KL}}
\newcommand{\B}{\mathrm{Beta}}

\newcommand{\barzv}{\bm{\bar{z}}}

\newtheorem{theorem}{Theorem}[section]
\newtheorem{lemma}[theorem]{Lemma}

\usepackage{color}

\usepackage[accepted]{icml2014}

\icmltitlerunning{Online Bayesian Passive-Aggressive Learning}

\begin{document}

\twocolumn[
\icmltitle{Online Bayesian Passive-Aggressive Learning}

% It is OKAY to include author information, even for blind
% submissions: the style file will automatically remove it for you
% unless you've provided the [accepted] option to the icml2013
% package.
\icmlauthor{Tianlin Shi}{stl501@gmail.com}
\icmladdress{Institute for Interdisciplinary Information Sciences,
            Tsinghua, Beijing}
\icmlauthor{Jun Zhu}{dcszj@mail.tsinghua.edu.cn}
\icmladdress{Department of Computer Science and Technology,
            Tsinghua, Beijing}

% You may provide any keywords that you
% find helpful for describing your paper; these are used to populate
% the "keywords" metadata in the PDF but will not be shown in the document
\icmlkeywords{bayesian, passive, aggressive, medlda, medhdp, machine learning, ICML}

\vskip 0.3in
]

\begin{abstract}
Online Passive-Aggressive (PA) learning is an effective framework for performing max-margin online learning. But the deterministic formulation and estimated single large-margin model could limit its capability in discovering descriptive structures underlying complex data. This paper presents online Bayesian Passive-Aggressive (BayesPA) learning, which subsumes the online PA and extends naturally to incorporate latent variables and perform nonparametric Bayesian inference, thus providing great flexibility for explorative analysis. We apply BayesPA to topic modeling and derive efficient online learning algorithms for max-margin topic models. We further develop nonparametric methods to resolve the number of topics. Experimental results on real datasets show that our approaches significantly improve time efficiency while maintaining comparable results with the batch counterparts.

%Existing inference paradigms for max-margin Bayesian learning require multi-passes through the dataset. In this paper, we present Bayesian Passive-Aggressive (PA) learning for supervised Bayesian models in the presence of online streaming data. Conceptually, our approach generalizes the online Passive-Aggressive algorithms by \cite{shalev2003online}, with new techniques to deal with latent variables. As case studies, we apply Bayesian PA learning to two maximum entropy discriminant models. Empirical experiments on real datasets show that our approach yield results comparable to batch counterparts in one pass, and significantly outperforms other online learning algorithms.
\end{abstract}

\section{Introduction}

% abstract knowledge -> models with structure -> Bayesian nonparametrics -> max-margin e.g. topic model.
% online learning.
% we propose bpal. -> generalize pa ->
% bpal applied to topic modeling.

Online learning is an effective way to deal with large-scale applications, especially applications with streaming data. Among the popular algorithms, online Passive-Aggressive (PA) learning~\cite{crammer2006pa} provides a generic framework for online large-margin learning, with many applications~\cite{McDonald2005parsing,chiang2008emnlp}. Though enjoying strong discriminative ability suitable for predictive tasks, existing online PA methods are formulated as a point estimate problem by optimizing some deterministic objective function. This may lead to some inconvenience. For example, a single large-margin model is often less than sufficient in describing complex data, such as those with rich underlying structures. %Moreover, the non-Bayesian formulation could restrict its marriage with Bayesian statistics for developing new methods.

On the other hand, Bayesian methods enjoy great flexibility in describing the possible underlying structures of complex data. Moreover, the recent progress on nonparametric Bayesian methods~\cite{hjort2010bayesian,teh2006hierarchical} further provides an increasingly important framework that allows the Bayesian models to have an unbounded model complexity, e.g., an infinite number of components in a mixture model~\cite{hjort2010bayesian} or an infinite number of units in a latent feature model~\cite{Griffiths:tr05}, and to adapt when the learning environment changes. For Bayesian models, one challenging problem is posterior inference, for which both variational and Monte Carlo methods can be too expensive to be applied to large-scale applications. To scale up Bayesian inference, much progress has been made on developing online variational Bayes~\cite{hoffman2010online,mimno2012sparse} and online Monte Carlo ~\cite{welling2012mc} methods. However, due to the generative nature, Bayesian models are lack of the discriminative ability of large-margin methods and usually less than sufficient in performing discriminative tasks.

Successful attempts have been made to bring large-margin learning and Bayesian methods together. For example, maximum entropy discrimination (MED)~\cite{Jaakkola:99} made a significant advance in conjoining max-margin learning and Bayesian generative models, mainly in the context of supervised learning and structured output prediction~\cite{Zhu:jmlr09}. Recently, much attention has been focused on generalizing MED to incorporate latent variables and perform nonparametric Bayesian inference, in many contexts including topic modeling~\cite{zhu2012medlda}, matrix factorization~\cite{Xu:NIPS12}, and multi-task learning~\cite{jebara2011mtl, zhu2011infinite}. However, posterior inference in such models remain a big challenge. It is desirable to develop efficient online algorithms for these Bayesian max-margin models.

To address the above problems of both the existing online PA algorithms and Bayesian max-margin models, this paper presents online Bayesian Passive-Aggressive (BayesPA) learning, a general framework of performing online learning for Bayesian max-margin models. %, with the great flexibility of Bayesian methods for explorative analysis as well as the strong discriminative ability of large-margin methods for predictive tasks.
We show that online BayesPA subsumes the standard online PA when the underlying model is linear and the parameter prior is Gaussian. We further show that another major significance of BayesPA is its natural generalization to incorporate latent variables and to perform nonparametric Bayesian inference, thus allowing online BayesPA to have the great flexibility of (nonparametric) Bayesian methods for explorative analysis as well as the strong discriminative ability of large-margin learning for predictive tasks. As concrete examples, we apply the theory of online BayesPA to topic modeling and derive efficient online learning algorithms for max-margin supervised topic models~\cite{zhu2012medlda}. We further develop efficient online learning algorithms for the nonparametric max-margin topic models, an extension of the nonparametric topic models~\cite{teh2006hierarchical, wang2011onlinealgo} for predictive tasks. Extensive empirical results on real data sets show significant improvements on time efficiency and maintenance of comparable results with the batch counterparts.

\section{Bayesian Passive-Aggressive Learning}\label{sec:bpal}

In this section, we present a general perspective on online max-margin Bayesian inference.

\subsection{Online PA Learning}\label{sec:pa}

The goal of online supervised learning is to minimize the cumulative loss for a certain prediction task from the sequentially arriving training samples. Online Passive-Aggressive (PA) algorithms \cite{crammer2006pa} achieve this goal by updating some parameterized model $\bm{w}$ (e.g., the weights of a linear SVM) in an online manner with the instantaneous losses from arriving data $\{\bm{x}_t\}_{t \geq 0}$ and corresponding responses $\{y_t\}_{t \geq 0}$. The losses $\ell_\epsilon(\bm{w}; \bm{x}_t, y_t)$, as they consider, could be the hinge loss $(\epsilon-y_t \wv^\top \xv_t)_+$ for binary classification or the $\epsilon$-insensitive loss $(|y_t- \wv^\top \xv_t|-\epsilon)_+$ for regression, where $\epsilon$ is a hyper-parameter and $(x)_+ = \max(0,x)$. The Passive-Aggressive update rule is then derived by defining the new weight $\bm{w}_{t+1}$ as the solution to the following optimization problem:
\begin{equation}\label{eq:pa}
\min_{\bm{w}}{\frac{1}{2} ||\bm{w}-\bm{w_{t}}||^2} ~~~ \text{s.t.:} ~ \ell_{\epsilon}(\bm{w}; \bm{x}_t, y_t) = 0.
\end{equation}
Intuitively, if $\bm{w_t}$ suffers no loss from the new data, i.e., $\ell_{\epsilon}(\bm{w}_t; \bm{x}_t, y_t) = 0$, the algorithm \emph{passively} assigns $\bm{w}_{t+1} = \bm{w}_t$; otherwise, it aggressively projects $\bm{w_t}$ to the feasible zone of parameter vectors that attain zero loss. With provable bounds, \cite{crammer2006pa} shows that online PA algorithms could achieve comparable results to the optimal classifier $\bm{w}^*$. In practice, in order to account for inseparable training samples, soft margin constraints are often adopted and the resulting learning problem is
\begin{equation}\label{eq:pa-soft}
\min_{\bm{w}}{\frac{1}{2} ||\bm{w}-\bm{w_{t}}||^2} + 2 c \ell_{\epsilon}(\bm{w}; \bm{x}_t, y_t),
\end{equation}
where $c$ is a positive regularization parameter. For problems~(\ref{eq:pa}) and (\ref{eq:pa-soft}) with samples arriving one at a time, closed-form solutions can be derived~\cite{crammer2006pa}.

\subsection{Online BayesPA Learning}
%We generalize online PA learning to the Bayesian setting.
Instead of updating a point estimate of $\bm{w}$, online Bayesian PA (BayesPA) sequentially infers a new posterior distribution $q_{t+1}(\bm{w})$, either parametric or nonparametric, on the arrival of new data $(\xv_t, y_t)$ by solving the following optimization problem:
%Classification problems are considered in this paper. Specifically, the following update rule is considered:
\setlength\arraycolsep{1pt} \begin{equation}\label{eq:onlinepa}
\begin{array}{rl}
\underset{q(\bm{w}) \in \mathcal{F}_t}{\operatorname{min}} & ~\text{KL}[q(\bm{w}) || q_{t}(\bm{w})]-\mathbb{E}_{q(\bm{w})}[\log p(\bm{x}_t | \bm{w})] \\
\text{s.t.:} &~~ \ell_\epsilon[q(\bm{w}); \bm{x}_t, y_t] = 0,
\end{array}
\end{equation}
where $\mathcal{F}_t$ is some distribution family, e.g., the probability simplex $\prob$. In other words, we find a posterior distribution $q_{t+1}(\bm{w})$ in the feasible zone that is not only close to $q_t(\bm{w})$ by the commonly used KL-divergence, but also has a high likelihood of explaining new data. %Notice that the likelihood term is peculiar to the Bayesian setting.
As a result, if Bayes' rule already gives the posterior distribution $q_{t+1}(\bm{w}) \propto q_{t}(\bm{w}) p(\bm{x}_t | \bm{w})$ that suffers no loss (i.e., $\ell_\epsilon = 0$), BayesPA \emph{passively} updates the posterior following just Bayes' rule; otherwise, BayesPA \emph{aggressively} projects the new posterior to the feasible zone of posteriors that attain zero loss. We should note that when no likelihood is defined (e.g., $p(\xv_t|\wv)$ is independent of $\wv$), BayesPA will passively set $q_{t+1}(\wv) = q_t(\wv)$ if $q_t(\wv)$ suffers no loss. We call it {\it non-likelihood} BayesPA.

In practical problems, the constraints in (\ref{eq:onlinepa}) could be unrealizable. To deal with such cases, we introduce the soft-margin version of BayesPA learning, which is equivalent to minimizing the objective function $\mathcal{L}(q(\bm{w}))$ in problem (\ref{eq:onlinepa}) with a regularization term \cite{cortes1995support}:
\setlength\arraycolsep{-3pt}\begin{eqnarray}\label{eq:onlinepa_reg}
&&q_{t+1}(\bm{w}) = \underset{q(\bm{w}) \in \mathcal{F}_t}{\operatorname{argmin}} ~\mathcal{L}(q(\bm{w})) + 2 c \ell_\epsilon(q(\bm{w}); \bm{x}_t, y_t).
\end{eqnarray}
%where $\mathcal{L}(q(\bm{w}))$ is the objective in problem~(\ref{eq:onlinepa}). %, and $c$ is a parameter controlling the extent of regularization.
For the max-margin classifiers that we focus on in this paper, two loss functions $\ell_\epsilon(q(\bm{w}); \bm{x}_t, y_t)$ are common --- the hinge loss of an \emph{averaging classifier} that makes predictions using the rule $\hat{y}_t = \textrm{sign} ~ \mathbb{E}_{q(\bm{w})}[\bm{w}^\top \bm{x}_t]$:
\setlength\arraycolsep{1pt}\begin{equation*}
%\label{eq:bayes_loss}
\ell_\epsilon^{Avg}[q(\bm{w}); \bm{x}_t, y_t] = (\epsilon-y_t \mathbb{E}_{q(\bm{w})}[\bm{w}^\top \bm{x}_t])_+
\end{equation*}
and the expected hinge loss of a \emph{Gibbs classifier} that randomly draws a classifier $\bm{w} \sim q(\bm{w})$ to make predictions using the rule $\hat{y}_t = \textrm{sign}~\bm{w}^\top \bm{x}_t$:
\begin{equation*}
%\label{eq:gibbs_loss}
\ell_\epsilon^{Gibbs}[q(\bm{w}); \bm{x}_t, y_t] = \mathbb{E}_{q(\bm{w})}[(\epsilon-y_t \bm{w}^\top \bm{x}_t)_+].
\end{equation*}
They are closely connected via the following lemma due to the convexity of the function $(x)_+$.
\begin{lemma}\label{lm:loss}
The expected hinge loss $\ell_\epsilon^{\text{Gibbs}}$ is an upper bound of the hinge loss $\ell_\epsilon^{\text{Avg}}$, that is, $\ell_\epsilon^{\text{Gibbs}} \geq \ell_\epsilon^{\text{Avg}}$.
\end{lemma}
%\begin{proof}
%From (\ref{eq:gibbs_loss}) we see that $\ell_\epsilon^{\text{Gibbs}} \geq \mathbb{E}_{q(\bm{w})} [\epsilon-y_t \bm{w}\top \bm{x}_t]$ and $\ell'_\epsilon \geq 0$. Therefore, $\ell_\epsilon^{\text{Gibbs}} \geq \max(0, \mathbb{E}_{q(\bm{w})} [\epsilon-y_t \bm{w}\top \bm{x}_t]) = \ell_\epsilon^{\text{Bayes}}$.
%\end{proof}
%Unlike PA algorithms where we have a closed-form update rule, it is unclear how one can efficiently solve (\ref{eq:onlinepa_reg}) with these two types of losses. In fact, Bayesian PA learning is a Bayesian inference problem with posterior regularization \cite{zhuregbayes}, for which exact solutions are known to be difficult. Nevertheless, approximate inference algorithms have been developed: variational inference for the margin loss   \cite{zhu2011infinite, zhu2012medlda}, and Gibbs sampling for the expected margin loss \cite{zhugibbs2013, zhu2013scalable}.
Before developing BayesPA learning for practical problems, we make several observations. %its connections to known algorithms.
%One is variational inference for the margin loss function \cite{zhu2011infinite, zhu2012medlda}, which poses strict mean-field assumptions on the posterior distribution. The other is Gibbs sampling for the expected margin loss \cite{zhugibbs2013, zhu2013scalable}, which is more accurate but merely gives a sample-based representation of the posterior in each round.
%PA Algorithms use Lagrange methods to solve the optimization and therefore attain simple additive rules. Unfortunately, the Bayesian PA Learning (\ref{eq:onlinepa})  is a \emph{regularzied} Bayesian inference problem with \emph{max-margin posterior constraints} \cite{zhuregbayes}, whose primal and dual problems are hard to solve in exact forms \cite{zhu2011infinite}. Nevertheless, we could use existing iterative algorithms \cite{zhu2009medlda} \cite{zhu2011infinite} to obtain approximate solutions efficiently.
\begin{lemma} \label{lm:subsume}
If $q_0(\bm{w}) = \mathcal{N}(0, I)$, $\mathcal{F}_t=\prob$ and we use $\ell_\epsilon^{Avg}$, the non-likelihood BayesPA subsumes the online PA.
\end{lemma}
This can be proved by induction. First, we can show that $q_t(\bm{w}) = \mathcal{N}(\bm{\mu}_t, I)$ is a normal distribution with an identity covariance matrix. Second, we can show that the posterior mean $\bm{\mu}_t$ is updated in the same way as in the online PA. We defer the detailed proof to Appendix A. %\ref{app:subsume}.
%\textbf{Connection to PA Algorithms:} The online PA algorithm \cite{crammer2006pa} is a special case of the BayesPA with an averaging prediction rule. To see this, we restrict the space of distribution families $\mathcal{F}$ to be the set of isotropic Gaussians: $\mathcal{F}  = \{\mathcal{N}(\bm{w}; \bm{w}_0, \sigma^2) | \bm{w}_0 \in \mathbb{R}^n\}$ for some fixed $\sigma \in \mathbb{R}$. For $q_{t+1}(\bm{w}) = \mathcal{N}(\bm{w}; \bm{w}_{t+1}, \sigma^2)$, Bayesian PA learning (\ref{eq:onlinepa_reg}) searches for optimal parameter $\bm{w}_{t+1}$. Since no likelihood is specified, we simply let $\bm{x}_t | \bm{w}$ be distributed uniformly. So
%\begin{equation}\label{eq:pa_again}
%\bm{w}_{t+1} =  \underset{\bm{w} \in \mathbb{R}^n}{\operatorname{argmin}}{\frac{1}{\sigma^2} ||\bm{w}-\bm{w}_t||^2}+c \ell_\epsilon(\bm{w}; \bm{x}_t, y_t),
%\end{equation}
%which is just (\ref{eq:pa}) with a different choice of $c$.
\begin{lemma} \label{lm:pagibbs}
If $\mathcal{F}_t=\prob$ and we use $\ell_\epsilon^{Gibbs}$, the update rule of online BayesPA is\\[-1cm]

\begin{equation}\label{eq:onlinepa_gibbs}
q_{t+1}(\wv) = \frac{q_{t}(\wv) p(\xv_t | \wv) e^{-2 c (\epsilon-y_t \wv^\top \xv_t)_+}}{\Gamma(\xv_t, y_t)},
\end{equation}\\[-1cm]

where $\Gamma(\xv_t, y_t)$ is the normalization constant.
\end{lemma}
%\textbf{Connection to Bayes Rule: }  In light of Zellner's fresh interpretation of optimal information processing \cite{zellner1988optimal}, Bayesian PA learning with expected margin loss could be described by Bayes rules. In other words, solving problem (\ref{eq:onlinepa_reg}) with expected margin loss $\ell'_\epsilon$, we get the normalized distribution,
Therefore, the posterior $q_t(\bm{w})$ in the previous round $t$ becomes a prior, while the newly observed data and its loss function provide a likelihood and an unnormalized pseudo-likelihood respectively. %Notice that the inference of (\ref{eq:onlinepa_gibbs}) could be made very efficient by exploiting data augmentation techniques~\cite{polson2011data}.

\textbf{Mini-Batches.} A useful technique to reduce the noise in data is the use of mini-batches. Suppose that we have a mini-batch of data points at time $t$ with an index set $B_t$, denoted as $\Xv_t = \{\xv_d\}_{d \in B_t}, \Yv_t = \{y_{d}\}_{d \in B_t}$. The Bayesian PA update equation for this mini-batch is simply,\\[-1cm]

\begin{equation*} %\label{eq:onlinepa_latent_batch}
q_{t+1}(\wv) = \underset{q \in \mathcal{F}_t}{\operatorname{argmin}}{~ \mathcal{L}(q(\wv)) +2 c \ell_\epsilon(q(\wv); \Xv_t, \Yv_t)},
\end{equation*}\\[-1cm]

where $\ell_\epsilon(q(\wv); \Xv_t, \Yv_t) = \sum_{d \in B_t}{\ell_\epsilon(q(\wv); \xv_d, y_d)}$.

\subsection{Learning with Latent Structures}\label{sec:pa_latent}

To expressively explain complex real-word data, Bayesian models with latent structures have been extensively developed. The latent structures could typically be characterized by two kinds of latent variables --- \emph{local latent variables} $\hv_d$ ($d \geq 0$) that characterize the hidden structures of each observed data $\xv_d$ and \emph{global variables} $\Mv$ that capture the common properties shared by all data.

The goal of Bayesian PA learning with latent structures is therefore to update the distribution of $\Mv$ as well as weights $\wv$ based on each incoming mini-batch $(\Xv_t, \Yv_t)$ and their corresponding latent variables $\Hv_t = \{\hv_d\}_{d \in B_t}$. Because of the uncertainty in $\Hv_t$, we extend BayesPA to infer the joint posterior distribution, $q_{t+1}(\wv, \Mv, \Hv_t)$, as solving\\[-1cm]

\setlength\arraycolsep{-3pt}\begin{eqnarray}\label{eq:onlinepa_latent}
&& \underset{q \in \mathcal{F}_t}{\operatorname{min}}{~\mathcal{L}(q(\wv, \Mv, \Hv_t)) + 2 c \ell_\epsilon(q(\wv, \Mv, \Hv_t); \Xv_t, \Yv_t)},
\end{eqnarray}\\[-1cm]

where $\mathcal{L}(q) \!=\!  \text{KL}[q ||  q_t(\wv, \Mv) p_0(\Hv_t)] -\mathbb{E}_q[\log p(\Xv_t | \wv, \\ \Mv, \Hv_t)]$ and $\ell_\epsilon(q; \Xv_t, \Yv_t)$ is some cumulative margin-loss on the min-batch data induced from some classifiers defined on the latent variables $\Hv_t$ and/or global variables $\Mv$. Both the averaging classifiers and Gibbs classifiers can be used as in the case without latent variables. We will present concrete examples in the next section.

%To see the connection of (\ref{eq:onlinepa_latent}) to (\ref{eq:onlinepa}), let the latent variables $\Hv_t$ be fully observed, which implies the space of $\Hv_t$ is collapsed to a point. In that case, (\ref{eq:onlinepa_latent}) and (\ref{eq:onlinepa}) become equivalent just by a substitution of global variables.

%  The data likelihood is
%
%\begin{equation}
%\Pr[\bm{X}_t | \bm{w}] = \int_{\bm{Z}_t}{\Pr[\bm{X}_t, \bm{Z}_t | \bm{w}] d \bm{Z}_t}
%\end{equation}¥
%
%for continuos variables $\bm{Z}_t$, and for discrete $\bm{Z}_t$ as well, if the integral is replaced by summation. On the other hand, the loss function $\ell_\epsilon(q(\bm{\mathcal{M}}, \bm{w}); \bm{Z}_t, y_t)$ is now defined with respect to latent variables.

Before diving into the details, we should note that in real online setting, only global variables are maintained in the bookkeeping, while the local information in the streaming data is forgotten. However, (\ref{eq:onlinepa_latent}) gives us a distribution of $(\wv, \Mv)$ that is coupled with the local variables $\Hv_t$. Although in some cases we can marginalize out the local variables $\Hv_t$, in general we would not obtain a closed-form posterior distribution $q_{t+1}(\wv, \Mv)$ for the next optimization round, especially in dealing with some involved models like MedLDA \cite{zhu2012medlda}. Therefore, we resort to approximation methods, e.g., by posing additional assumptions about $q(\wv, \Mv, \Hv_t)$ such as the mean-field assumption, $q(\wv, \Mv, \Hv_t) = q(\wv) q(\Mv) q(\Hv_t)$. Then, we can solve the problem via an iterative procedure and use the optimal distribution $q^*(\wv) q^*(\Mv)$ as $q_{t+1}(\wv, \Mv)$. More details will be provided in next sections.

\section{Online Max-Margin Topic Models}\label{sec:otm}

We apply the theory of online BayesPA to topic modeling and develop online learning algorithms for max-margin topic models. We also present a nonparametric generalization to resolve the number of topics in the next section.

%Before touching the case design of Bayesian PA learning, we introduce two max-margin topic models: one with finite latent representations -- maximum entropy discrimination Latent Dirichlet Allocation (MedLDA), and the other with infinite latent structure -- maximum entropy discriminant Hierarchical Dirichlet Process (MedHDP).

\subsection{Batch MedLDA} \label{sec:medlda}

A max-margin topic model consists of a latent Dirichlet allocation (LDA)~\cite{blei2003lda} model for describing the underlying topic representations and a max-margin classifier for predicting responses. Specifically, LDA is a hierarchical Bayesian model that treats each document as an admixture of topics, $\Phiv = \{\phiv_k\}_{k=1}^K$, where each topic $\bm{\phi}_k$ is a multinomial distribution over a $W$-word vocabulary. Let $\thetav$ denote the mixing proportions. The generative process of document $d$ is described as
\setlength\arraycolsep{1pt} \begin{eqnarray}
\thetav_d \sim && \text{Dir}(\bm{\alpha}), \nonumber \\
z_{di} \sim \text{Mult}(\thetav_d),~x_{di} &&\sim \text{Mult}(\phiv_{z_{di}}),~\forall i \in [n_d] \nonumber
\end{eqnarray}
%\begin{itemize}
%\item draw a topic portion $\thetav_d \sim \text{Dir}(\bm{\alpha})$;
%\item for word $i$ $(0 \leq i < n_d)$ in document $d$:
%\begin{itemize}
%\item draw topic assignments $z_{di} \sim \text{Mult}(\thetav_d)$.
%\item draw observed word $x_{di} \sim \text{Mult}(\phiv_{z_{di}})$.
%\end{itemize}
%\end{itemize}
where $z_{di}$ is a topic assignment variable and $\text{Mult}(\cdot)$ is a multinomial distribution. For Bayesian LDA, the topics are drawn from a Dirichlet distribution, i.e., $\bm{\phi}_k \sim \text{Dir}(\bm{\gamma})$.

Given a document set $\Xv = \{\xv_d\}_{d=1}^D$. Let $\bm{Z} = \{\bm{z}_d\}_{d=1}^D$ and $\Thetav = \{\bm{\theta}_d\}_{d=1}^{D}$. LDA infers the posterior distribution $p( \bm{\Phi}, \bm{\Theta}, \bm{Z} | \bm{X}) \propto p_0(\bm{\Phi}, \bm{\Theta}, \bm{Z}) p(\bm{X} | \bm{Z}, \bm{\Phi})$ via Bayes' rule. From a variational point of view, the Bayes posterior is equivalent to the solution of the optimization problem:
\begin{equation*}%\label{eq:ldazellner}
\min\limits_{q \in \mathcal{P}}~\KL[q(\Phiv, \Thetav, \Zv) || p(\Phiv, \Thetav, \Zv | \Xv) ]. %p_0(\bm{\Phi}, \bm{\Theta}, \bm{Z})]-\mathbb{E}_q[\log p(\bm{X}|\bm{Z}, \bm{\Phi})]}.
\end{equation*}
The advantage of the variational formulation of Bayesian inference lies in the convenience of posing restrictions on the post-data distribution with a regularization term. For supervised topic models \cite{blei2010supervised,zhu2012medlda}, such a regularization term could be a loss function of a prediction model $\bm{w}$ on the data $\Xv = \{\xv_d\}_{d=1}^D$ and response signals $\Yv = \{y_d\}_{d=1}^D$. As a regularized Bayesian (RegBayes) model \cite{jiang2012monte}, MedLDA infers a distribution of the latent variables $\Zv$ as well as classification weights $\wv$ by solving the problem:
\setlength\arraycolsep{-2pt}\begin{eqnarray*}
&& \min\limits_{q \in \mathcal{P}}  ~\mathcal{L}( q(\wv, \Phiv, \Thetav, \Zv) ) + 2 c \sum\limits_{d=1}^{D}{\ell_{\epsilon}(q(\wv, \zv_d); \xv_d, y_d)},
\end{eqnarray*}
where \small $\mathcal{L}( q(\wv, \Phiv, \Thetav, \Zv) ) = \KL[q(\wv, \Phiv, \Thetav, \Zv) || p(\wv, \Phiv, \Thetav,\\ \Zv | \Xv)]$ \normalsize. To specify the loss function, a linear discriminant function needs to be defined with respect to $\bm{w}$ and $\bm{z}_d$\\[-1cm]

\setlength\arraycolsep{1pt}\begin{equation}\label{eq:disc-func-latent}
f(\wv, \zv_d) = \wv^\top \bar{\zv}_d,
\end{equation}\\[-1cm]

where $\bar{\zv}_{dk} = \frac{1}{n_d} \sum_i{\mathbb{I}[z_{di} = k]}$ is the average topic assignments of the words in document $d$. Based on the discriminant function, both averaging classifiers with the hinge loss
\begin{equation} \label{eq:batch_bayes_loss}
\ell^{Avg}_{\epsilon}(q(\wv, \zv_d); \xv_d, y_d) = (\epsilon - y_d \mathbb{E}_q[f(\bm{w}, \bm{z}_d)])_+,
\end{equation}
and Gibbs classifiers with the expected hinge loss
\begin{equation}  \label{eq:batch_gibbs_loss}
\ell^{Gibbs}_{\epsilon}(q(\wv, \zv_d); \xv_d, y_d) = \mathbb{E}_q[(\epsilon - y_d f(\bm{w}, \bm{z}_d))_+],
\end{equation}
have been proposed, with extensive comparisons reported in~\cite{zhugibbs2013} using batch learning algorithms.

\subsection{Online MedLDA}\label{sec:pamedlda}

To apply the online BayesPA, we have the global variables $\Mv = \Phiv$ and local variables $\Hv_t = (\Thetav_t, \Zv_t)$. We consider Gibbs MedLDA because as shown in~\cite{zhugibbs2013} it admits efficient inference algorithms by exploring data augmentation. Specifically, let $\zeta_d = \epsilon - y_d f(\wv, \zv_d)$ and $\psi(y_d | \zv_d, \wv) = e^{-2c (\zeta_d)_+}$. Then in light of Lemma \ref{lm:pagibbs}, the optimal solution to problem~(\ref{eq:onlinepa_latent}), $q_{t+1}(\wv, \Mv, \Hv_t)$,  is\\[-1cm]

\begin{eqnarray*}
%\label{eq:online-solution}
\frac{ q_t(\wv, \Mv) p_0(\Hv_t)  p(\Xv_t | \Hv_t, \Mv) \psi(\Yv_t | \Hv_t, \wv) }{ \Gamma( \Xv_t, \Yv_t ) },
\end{eqnarray*}\\[-1cm]

where \small $\psi(\Yv_t | \Hv_t, \wv) = \prod_{d \in B_t} \psi(y_d | \hv_d, \wv)$ \normalsize and \small $\Gamma(\Xv_t, \Yv_t)$ \normalsize is a normalization constant. To potentially improve the inference accuracy, we first integrate out the local variables $\Thetav_t$ by the conjugacy between a Dirichlet prior and a multinomial likelihood~\cite{griffiths2004finding,teh2006collapsed}. Then we have the local variables $\Hv_t = \Zv_t$. By the equality~\cite{zhugibbs2013}:\\[-.9cm]

%The philosophy behind the choice of Gibbs models, instead of their Bayes counterparts, is two-fold: first, by Lemma \ref{lm:loss}, the expected margin loss is an upper bound of the margin loss. Second and most importantly, Lemma 2 in \cite{zhugibbs2013} shows that a surprisingly simple form of online BayesPA learning could be derived.
%\begin{lemma}\label{lm: sm}
%\textbf{(Scale Mixture Representation)} The following equality holds true,
\begin{equation} \label{eq:scalemix}%e^{-2c (\zeta_d)_{+}}
\psi(y_d | \zv_d, \wv) = \int_{0}^{\infty}{\psi(y_d, \lambda_d | \bm{z}_d, \bm{w})d\lambda_d},
\end{equation}\\[-1cm]

where $\psi(y_d, \lambda_d | \zv_d, \wv) = (2\pi\lambda_d)^{-1/2}\exp(-\frac{(\lambda_d+c\zeta_d)^2}{2\lambda_d})$, the collapsed posterior $q_{t+1}(\wv, \Phiv, \Zv_t)$ is a marginal distribution of $q_{t+1}(\wv, \Phiv, \Zv_t, \lambdav_t)$, which equals to\\[-1cm]

\setlength\arraycolsep{-3pt}  \begin{eqnarray*}\label{eq:online-solution-augmented}
&& \frac{ p_0(\Zv_t) q_t(\wv, \Phiv) p(\Xv_t | \Zv_t, \Phiv) \psi(\Yv_t, \lambdav_t | \Zv_t, \wv) }{ \Gamma( \Xv_t, \Yv_t ) },
\end{eqnarray*}\\[-1cm]

where \small $\psi(\Yv_t, \lambdav_t | \Zv_t, \wv) \!=\! \prod_{d \in B_t} \psi(y_d, \lambda_d | \zv_d, \wv)$ \normalsize and $\lambdav_t = \{\lambda_d\}_{d\in B_t}$ are augmented variables, which are also locally associated with individual documents. In fact, the augmented distribution $q_{t+1}(\wv, \Phiv, \Zv_t, \lambdav_t)$ is the solution to
%be attained via a Bayes rule with likelihood $\prod_{d \in B_t}p_t(\bm{x}_{d} | \bm{\mathcal{M}}, \bm{w}, \bm{z}_{d})$ and pseudo-likelihood $\prod_{d \in B_t}\psi(y_{d}, \lambda_{d} | \bm{z}_{d}, \bm{w})$, which is equivalent to solving
the problem:\\[-1cm]

\setlength\arraycolsep{-4pt} \begin{eqnarray} \label{eq:onlinepa_augmented}
&& \underset{q \in \mathcal{P}}{\operatorname{min}}{~\mathcal{L}(q(\wv, \Phiv, \Zv_t, \lambdav_t))- \ep_q[\log \psi(\Yv_t, \lambdav_t | \Zv_t, \wv)]},
\end{eqnarray}\\[-1cm]

where $\mathcal{L}(q) = \KL[ q(\wv, \Phiv, \Zv_t, \lambdav_t) \Vert q_t(\wv, \Phiv)  p_0(\Zv_t)]- \\ \ep_q[\log p(\Xv_t|\Zv_t, \Phiv) ]$. We can show that this objective is an upper bound of that in the original problem~(\ref{eq:onlinepa_latent}). See Appendix B for details. %in \ref{app:upperbound}

With the mild mean-field assumption that $q(\wv, \Phiv, \Zv_t, \lambdav_t) \\= q(\wv)q(\Phiv)q(\Zv_t, \lambdav_t)$, we can solve~(\ref{eq:onlinepa_augmented}) via an iterative procedure that alternately updates each factor distribution \cite{jordan1998introduction}, as detailed below.

{\bf Global Update:} By fixing the distribution of local variables, $q(\Zv_t, \lambdav_t)$, and ignoring irrelevant variables, we have the mean-field update equations:
%the optimal global distributions $q(\bm{\Phi})$ and $q(\bm{\eta})$ are decoupled. For prior distributions $q_t(\Phiv) \sim \text{Dir}(\Delta_{k1}^{t}, ..., \Delta_{kW}^{t}), q_t(\wv) \sim \mathcal{N}(\wv; \muv^t, \Sigma^t)$, the optimal solutions obey the following closed forms \cite{bishop2006}:
\setlength\arraycolsep{0pt} \begin{eqnarray}
&& q(\bm{\Phi}_k) \propto q_t(\Phiv_k) \exp( \ep_{q(\Zv_t)} [\log p_0(\Zv_t) p(\Xv|\Zv_t, \Phiv) ]  ), ~ \forall~  k \label{eq:lda_phi} \nonumber \\
&& q(\bm{w}) \propto  q_t(\wv) \exp(\ep_{q(\Zv_t,\lambdav_t)}[\log p_0(\Zv_t) \psi( \Yv_t , \lambdav_t | \Zv_t, \wv) ] ) . \label{eq:lda_weight} \nonumber
\end{eqnarray}
If initially $q_0(\Phiv_k) = \text{Dir}(\Delta_{k1}^{0}, ..., \Delta_{kW}^{0})$ and $q_0(\wv) = \mathcal{N}(\wv; \muv^0, \Sigmav^0)$, by induction we can show that the inferred distributions in each round has a closed form, namely, $q_t(\Phiv_k) = \text{Dir}(\Delta_{k1}^{t}, ..., \Delta_{kW}^{t})$ and $q_t(\wv) = \mathcal{N}(\wv; \muv^t, \Sigmav^t)$. For the above update equations, we have
\setlength\arraycolsep{1pt} \begin{equation}\label{eq:lda_phi2}
q(\bm{\Phi}_k) = \text{Dir}(\Delta_{k1}^*, ..., \Delta_{kW}^*),
\end{equation}
where $\Delta_{kw}^* = \Delta_{kw}^t + \sum_{d \in B_t}\sum_{i \in [n_d]}{ \gamma_{di}^k \cdot \mathbb{I}[x_{di} = w]}$ for all words $w$ and $\gamma_{di}^k = \ep_{q(\zv_d)} \mathbb{I}[z_{di} = k]$ is the probability of assigning word $x_{di}$ to topic $k$, and
\setlength\arraycolsep{1pt} \begin{equation}\label{eq:lda_weight2}
q(\wv) = \mathcal{N}(\bm{w}; \bm{\mu}^*, \Sigmav^*),
\end{equation}
where the posterior paramters are computed as $(\Sigmav^*)^{-1}  = (\Sigmav^t)^{-1} + c^2 \sum_{d \in B_t}\ep_{q(\zv_d,\lambda_d)}[\lambda_d^{-1} \bar{\zv}_d \bar{\zv}_d^\top ]$ and $\muv^* = \Sigmav^* (\Sigmav^t)^{-1} \muv^t+\Sigmav^* \cdot c\sum_{d \in B_t}\ep_{q(\zv_d, \lambda_d)}[y_d (1 + c\epsilon \lambda_d^{-1}) \bar{\zv}_d]$. %, respectively.

{\bf Local Update:} Given the distribution of global variables, $q(\Phiv,\wv)$, the mean-field update equation for $(\Zv_t, \lambdav_t)$ is
 \begin{eqnarray} \label{eq:pamedlda_zlambda}
 %\propto & p_0(\Zv_t) \exp(\ep_{q(\Phiv, \wv)} [\log p(\Xv_t | \Zv_t, \Phiv) \psi(\Yv_t, \lambdav_t | \Zv_t, \wv)]) \\\\
& q(\Zv_t, \lambdav_t) \propto & p_0(\Zv_t) \prod\limits_{d \in B_t} \frac{1}{\sqrt{2 \pi \lambda_d}}\exp\Big(\nonumber \sum\limits_{i \in [n_d]} \Lambda_{z_{di},x_{di}}\\
&& ~~~~~~~~~~~~~~~~~~   - \ep_{q(\Phiv,\wv)}[ \frac{(\lambda_d+c\zeta_d)^2}{2\lambda_d}] \Big),
\end{eqnarray}
where $\Lambda_{z_{di}, x_{di}} = \ep_{q(\Phi)}[\log(\Phi_{z_{di}, x_{di}})] = \Psi(\Delta_{z_{di}, x_{di}}^*)-\Psi(\sum_{w}{\Delta_{z_{di}, w}^*})$ and $\Psi(\cdot)$ is the digamma function, due to the distribution in~(\ref{eq:lda_phi2}). But it is impossible to evaluate the expectation in the global update using (\ref{eq:pamedlda_zlambda}) because of the huge number of configurations for $(\bm{Z}_t, \bm{\lambda}_t)$. As a result, we turn to Gibbs sampling and estimate the required expectations using multiple empirical samples. This hybrid strategy has shown promising performance for LDA~\cite{mimno2012sparse}. Specifically, the conditional distributions used in the Gibbs sampling are as follows:

\textbf{For $\Zv_t$:} By canceling out common factors, the conditional distribution of one variable $z_{di}$ given $\bm{Z}_{t}^{\neg di}$ and $\lambdav_t$ is
\setlength\arraycolsep{0pt}
\begin{equation}\label{eq:lda_sample_z}
\begin{array}{rl}
&\!q(z_{di} \!=\!  k| \Zv_t^{\neg di}, \lambdav_t) \!\propto\!  (\alpha \!+\! C_{dk}^{\neg di})  \! %\\ %(\Psi(\Delta_{k x_{di}}^*)-\Psi(\sum_w \Delta_{k w}^*))  \\
\exp\!\Big(\frac{c y_d(c\epsilon + \lambda_d) \mu_k^*}{n_d \lambda_d} \\
&~~~~~~~ + \Lambda_{k,x_{di}} - \frac{c^2 (\mu_{k}^{*2}+\Sigma_{kk}^* + 2 (\mu_k^* \muv^* + \Sigmav_{\cdot, k}^* )^\top \bm{C}_{d}^{\neg di} ) }{2 n_d^2 \lambda_d}\Big),
\end{array}
\end{equation}
where $\Sigmav_{\cdot, k}^*$ is the $k$-th column of $\Sigmav^*$, $\bm{C}_d^{\neg di}$ is a vector with the $k$-th entry being the number of words in document $d$ (except the $i$-th word) that are assigned to topic $k$.

\textbf{For $\lambdav_t$:} Let $\bar{\zeta}_d = \epsilon-y_d \bar{\bm{z}}_d^\top \bm{\mu}^*$. The conditional distribution of each variable $\lambda_d$ given $\Zv_t$ is\\[-1cm]

\begin{equation}\label{eq:lda_sample_lambda}
\begin{array}{rl}
q(\lambda_d | \Zv_t) \propto & \frac{1}{\sqrt{2 \pi \lambda_d}} \exp\left( - \frac{c^2 \bar{\bm{z}}_d^\top \Sigmav^* \bar{\bm{z}}_d + (\lambda_d+c \bar{\zeta}_d )^2}{2\lambda_d} \right) \\\\
= & \mathcal{GIG}\left( \lambda_d; \frac{1}{2}, 1, c^2 (\bar{\zeta}_d^2+\bar{\bm{z}}_d^\top\Sigmav^*\bar{\bm{z}}_d) \right),
\end{array}
\end{equation}\\[-1cm]

a generalized inverse gaussian distribution \cite{devroye1986sample}. Therefore, $\lambda_d^{-1}$ follows an inverse gaussian distribution $\mathcal{IG}(\lambda_d^{-1}; \frac{1}{c\sqrt{\bar{\zeta}_d^2+\bar{\bm{z}}_d^\top\Sigmav^* \bar{\bm{z}}_d}}, 1)$, from which we can draw a sample in constant time~\citep{Michael:IG76}.

For training, we run the global and local updates alternately until convergence at each round of PA optimization, as outlined in Alg.~\ref{algo:pamedlda}. To make predictions on testing data, we then draw one sample of $\hat{\bm{w}}$ as the classification weight and apply the prediction rule. The inference of $\barzv$ for testing documents is the same as in ~\cite{zhugibbs2013}.

\begin{algorithm}[t]
\caption{\small{Online MedLDA}}
\label{algo:pamedlda}
\begin{algorithmic}[1]
\STATE Let $q_0(\wv) = \mathcal{N}(0; v^2 I), q_0(\phiv_k) = \text{Dir}(\gamma), ~\forall~ k$.
\FOR{$t = 0 \to \infty$}
\STATE Set $q(\Phiv, \wv) = q_t(\Phiv, \wv)$. Initialize $\bm{Z}_t$.
\FOR{$i = 1 \to \mathcal{I}$}
\STATE Draw samples $\{\Zv_t^{(j)}, \lambdav_t^{(j)}\}_{j=1}^{\mathcal{J}}$ from  (\ref{eq:lda_sample_z}, \ref{eq:lda_sample_lambda}).
\STATE Discard the first $\beta$ burn-in samples ($\beta < \mathcal{J}$).
\STATE Use the rest $\mathcal{J}-\beta$ samples to update $q(\Phiv, \wv)$ following (\ref{eq:lda_phi2}, \ref{eq:lda_weight2}).
\ENDFOR
\STATE Set $q_{t+1}(\Phiv, \wv) = q(\Phiv, \wv)$.
\ENDFOR
\end{algorithmic}
\end{algorithm}

\section{Online Nonparametric MedLDA}\label{sec:ohdp}

We present online nonparametric MedLDA for resolving the unknown number of topics, based on the theory of hierarchical Dirichlet process (HDP) \cite{teh2006hierarchical}.

\subsection{Batch MedHDP} \label{sec:medhdp}

HDP provides an extension to LDA that allows for a nonparametric inference of the unknown topic numbers. The generative process of HDP can be summarized using a stick-breaking construction~\cite{wang2012truncation}, where the stick lengths $\bm{\pi} = \{\pi_k\}_{k = 1}^{\infty}$ are generated as:\\[-1cm]

\begin{equation*}
\begin{array}{l}
\pi_k = \bar{\pi}_k \prod\limits_{i < k}(1-\bar{\pi}_{i}), ~ \bar{\pi}_k \sim \text{Beta}(1,\gamma),~ \textrm{for}~k=1, ..., \infty,
\end{array}
\end{equation*}\\[-1cm]

and the topic mixing proportions are generated as $\thetav_d \sim \text{Dir}(\alpha \piv),~\textrm{for}~d = 1,...,D$. Each topic $\bm{\phi}_k$ is a sample from a Dirichlet base distribution, i.e., $\bm{\phi}_k \sim \text{Dir}(\bm{\eta})$. After we get the topic mixing proportions $\thetav_d$, the generation of words is the same as in the standard LDA.

%The goal of HDP is to infer the posterior distribution   $p(\Phiv, \Thetav, \barpiv, \Zv | \Xv)$ from data $\bm{X}$, which is the solution to the following optimization problem:
%\begin{equation}\label{eq:hdpzellner}
%\min\limits_{q \in \mathcal{P}} ~\KL[q(\Phiv, \Thetav, \barpiv, \Zv) || p(\Phiv, \Thetav, \barpiv, \Zv | \Xv)]. %-\mathbb{E}_q[\log p(\bm{X}|\bm{Z}, \bm{\Phi})]}.
%\end{equation}
To augment the HDP topic model for predictive tasks, we introduce a classifier $\wv$ and define the linear discriminant function in the same form as~(\ref{eq:disc-func-latent}),
%\begin{equation}\label{eq:discri_infinite}
%f(\wv, \zv_d) = \wv^\top \bar{\zv}_d,   %\sum\limits_{k=1}^{\infty}{w_k  \bar{z}_k}.
%\end{equation}
where we should note that since the number of words in a document is finite, the average topic assignment vector $\bar{\zv}_{d}$ has only a finite number of non-zero elements. Therefore, the dot product in (\ref{eq:disc-func-latent}) is in fact finite. Let $\bm{\bar{\pi}} = \{\bar{\pi}_k\}_{k=1}^{\infty}$. We define MedHDP as solving the following problem to infer the joint posterior $q(\wv, \barpiv, \Phiv, \Thetav,  \Zv)$\footnote{Given $\barpiv$, $\piv$ can be computed via the stick breaking process.}:\\[-1.1cm]

\setlength\arraycolsep{-2pt} \begin{eqnarray*}
&& \min\limits_{q \in \mathcal{P}}  {\mathcal{L}(q(\wv, \barpiv, \Phiv, \Thetav, \Zv))}+ 2 c \sum\limits_{d=1}^{D}{\ell_{\epsilon}(q(\wv, \zv_d); \xv_d, \yv_d)},
\end{eqnarray*}\\[-1cm]

where \small $\mathcal{L}( q(\wv, \Phiv, \barpiv, \Thetav, \Zv) ) = \KL[q(\wv, \Phiv, \barpiv, \Thetav, \Zv) || p(\wv, \\ \barpiv, \Phiv, \Thetav,  \Zv | \Xv)]$\normalsize, and the loss function could be either (\ref{eq:batch_bayes_loss}) or (\ref{eq:batch_gibbs_loss}), leading to the MedHDP topic models with either averaging or Gibbs classifiers.

%Yet, given the infinite space of MedHDP, it is unclear how one could devise a truncation-free \cite{wang2012truncation, wang2011online} inference algorithm. This issue would be further discussed in section \ref{sec:pamedhdp}.

%We introduce ideas of regularized Bayesian inference into HDP and propose a new infinite supervised topic model: max entropy discrimination HDP (MedHDP).

\subsection{Online MedHDP}\label{sec:pamedhdp}

To apply the online BayesPA, we have the global variables $\Mv = (\barpiv, \Phiv)$, and the local variables $\Hv_t = (\Thetav_t, \Zv_t)$. We again focus on the expected hinge loss~(\ref{eq:batch_gibbs_loss}) in this paper. As in online MedLDA, we marginalize out $\Thetav_t$ and adopt the same data augmentation technique with the augmented variables $\lambdav_t$. Furthermore, to simplify the sampling scheme, we introduce auxiliary latent variables $\Sv_t = \{\sv_d\}_{d \in B_t}$, where $s_{dk}$  represents the number of occupied tables serving dish $k$ in a Chinese Restaurant Process~\cite{teh2006hierarchical, wang2012truncation}. By definition, we have $p(\Zv_t, \Sv_t | \barpiv) =  \prod_{d \in B_t} p(\sv_d, \zv_d | \barpiv)$  and\\[-1cm]

\begin{equation} \label{eq:joint_zs}
p(\sv_d, \zv_d | \barpiv) \propto  \prod_{k = 1}^{\infty}{S( n_d \bar{z}_{dk}, s_{dk}) (\alpha \pi_k)^{s_{dk}}},
\end{equation}\\[-1cm]

where  $S(a,b)$ are unsigned Stirling numbers of the first kind \cite{antoniak1974mixtures}. It is not hard to verify that $p(\zv_d | \barpiv) = \sum_{\sv_d}{p(\sv_d, \zv_d | \barpiv}$). Therefore, we have local variables $\Hv_t = (\Zv_t, \Sv_t, \lambdav_t)$, and the target collapsed posterior $q_{t+1}(\wv, \barpiv, \Phiv, \Zv_t, \lambdav_t)$ is the marginal distribution of $q_{t+1}(\wv, \barpiv, \Phiv, \Hv_t)$, which is the solution of the problem:\\[-1cm]

\setlength\arraycolsep{-4pt}  \begin{eqnarray} \label{eq:onlinemedhdp_augmented}
&& \underset{q \in \mathcal{F}_t}{\operatorname{min}}{~\mathcal{L}(q(\wv, \barpiv, \Phiv, \Hv_t)\!)
\!-\!\ep_q[\log \psi(\Yv_t, \lambdav_t | \Zv_t, \wv)]},
\end{eqnarray}\\[-1cm]

%\setlength\arraycolsep{-5pt}  \begin{eqnarray}\label{eq:post_medhdp}
%&& \frac{ p(\Zv_t, \Sv_t | \barpiv) q_t(\Phiv, \barpiv, \wv) p(\Xv_t | \Zv_t, \Phiv) \psi(\Yv_t, \lambdav_t | \Zv_t, \wv) }{ \Gamma( \Xv_t, \Yv_t ) },
%\end{eqnarray}
where \small $\mathcal{L}(q) = \KL[q || q_t(\wv, \barpiv, \Phiv) p(\Zv_t, \Sv_t | \barpiv) p(\Xv_t | \Zv_t, \Phiv)]$ \normalsize. As in online MedLDA, we solve (\ref{eq:onlinemedhdp_augmented}) via an iterative procedure detailed below.

\iffalse
Then we can apply online BayesPA rule (\ref{eq:onlinepa_augmented}) by maximizing the ELBO:
\begin{equation}\label{eq:pamedhdp_eblo}
\underset{q \in \mathcal{F}}{\operatorname{max}}{~\mathbb{E}_q[\log p(\bm{X}_t, \bm{Y}_t , \bm{\Phi}, \bm{\bar{\pi}}, \bm{Z}_t, \bm{S}_t, \bm{\lambda}_t)]+\mathbb{H}[q]}
\end{equation}
where
\begin{equation*}
\begin{array}{rl}
& \log p(\bm{X}_t, \bm{Y}_t , \bm{\Phi}, \bm{\bar{\pi}}, \bm{Z}_t, \bm{S}_t, \bm{\lambda}_t)  \\\\
= & \sum\limits_{d \in B_t}{\mathbb{E}_{q}\log [ \prod_{i}{\phi_{z_{di}, x_{di}}}]} +\mathbb{E}_q\log p(\bm{s}_{d} , \bm{\bar{z}}_{d} | \bm{\pi}) \\
 & ~ + \sum\limits_{d \in B_t}{\mathbb{E}_{q}[\log \psi(y_d, \lambda_d | \bm{z}_d, \bm{w})]} \\
 &  ~ + \mathbb{E}_q{\log q_t(\bm{\Phi)}} + \mathbb{E}_q{\log q_t(\bm{w)}} + \mathbb{E}_{q}{\log q_t{(\bm{\pi})}}.
\end{array}
\end{equation*}
\fi

\textbf{Global Update: }  By fixing the distribution of local variables, $q(\Zv_t, \Sv_t, \lambdav_t)$, and ignoring the irrelevant terms, we have the mean-field update equations for $\Phiv$ and $\wv$, the same as in (\ref{eq:lda_phi2}) and (\ref{eq:lda_weight2}), while for $\bar{\piv}$, we have
%Assume prior distributions $q_t(\Phiv) \sim \text{Dir}(\Delta_{k1}^{t}, ..., \Delta_{kW}^{t}), q_t(\wv) \sim \mathcal{N}(\wv; \muv^t, \Sigma^t), q_t(\bar{\pi}_k) = \text{Beta}(u_k^t, v_k^t)$. , the global distribution of $\bm{\Phi}$, $\bm{w}$ shares the same form with (\ref{eq:lda_phi}) and (\ref{eq:lda_weight}), while for $\bm{\bar{\pi}}$,
\begin{equation} \label{eq:hdp_beta}
q(\bar{\pi}_k) \propto q_t(\bar{\pi}_k) \prod_{d \in B_t} \exp( \ep_{q(\hv_d)} [ \log p(\sv_d, \zv_d | \barpiv) ]).
%bm{X}_t, \bm{Y}_t , \bm{\Phi}, \bm{\bar{\pi}}, \bm{w}, \bm{Z}_t, \bm{S}_t, \bm{\lambda}_t)]  ) \\
%= & \text{Beta}(u_{k}, v_{k})
\end{equation}
By induction, we can show that $q_t(\bar{\pi}_k) = \text{Beta}(u_k^t, v_k^t)$, a Beta distribution at each step, and the update equation is
\begin{equation}
q(\bar{\pi}_k) = \text{Beta}(u_k^*, v_k^*),
\end{equation}
where $u_k^* = u_k^t + \sum_{d\in B_t} \ep_{q(\sv_d)}{[s_{dk}]}$ and $v_k^* = v_k^t + \sum_{d \in B_t} \ep_{q(\sv_d)}{[\sum_{j > k}{s_{dj}}]}$ for $k = \{1,2,...\}$. Since $\bm{Z}_t$ contains only finite number of discrete variables, we only need to maintain and update the global distribution for a finite number of topics.

\textbf{Local Update:} Fixing the global distribution $q(\bm{w}, \barpiv, \Phiv)$, we get the mean-field update equation for $(\Zv_t, \Sv_t, \lambdav_t)$:
\begin{equation} \label{eq:pamedhdp_zlambdas}
q(\bm{Z}_t, \bm{S}_t, \bm{\lambda}_t)  \propto \tilde{q}(\bm{Z}_t, \bm{S}_t) \tilde{q}(\bm{Z}_t, \bm{\lambda}_t)
\end{equation}
where $\tilde{q}(\bm{Z}_t, \bm{S}_t) = \exp( \mathbb{E}_{q(\bm{\Phi)} q(\bm{\bar{\pi}})}[\log p( \bm{X}_t | \bm{\Phi},  \bm{Z}_t)+\log p(\bm{Z}_t, \bm{S}_t | \bm{\bar{\pi}})])$ and $\tilde{q}(\bm{Z}_t, \bm{\lambda}_t) = \exp( \mathbb{E}_{q(\bm{w})}[\log \psi(\bm{Y}_t, \\\bm{\lambda}_t|\bm{w}, \bm{Z}_t)])$. To overcome the the potentially unbounded latent space, we take the ideas from~\cite{wang2012truncation} and adopt an approximation for $\tilde{q}(\Zv_t, \Sv_t)$:
\begin{equation} \label{eq:hdp_approx}
\tilde{q}(\bm{Z}_t, \bm{S}_t) \approx \mathbb{E}_{q(\bm{\Phi)} q(\bm{\bar{\pi}})}[p(\bm{X} | \bm{\Phi}, \bm{Z}_t) p(\bm{Z}_t, \bm{S}_t | \bm{\bar{\pi}})].
\end{equation}
Instead of marginalizing out $\barpiv$ in (\ref{eq:hdp_approx}), which is analytically difficult, we sample $\barpiv$ jointly with $(\Zv_t, \Sv_t, \lambdav_t)$. This leads to the following Gibbs sampling scheme:

\textbf{For $\bm{Z}_t$: } Let $K$ be the current inferred number of topics. The conditional distribution of one variable $z_{di}$ given $\bm{Z}_{t}^{\neg di}$, $\lambdav_t$ and $\barpiv$ can be derived from (\ref{eq:pamedhdp_zlambdas}) with $\sv_d$ marginalized out for convenience:
\setlength\arraycolsep{-3pt} \begin{equation*}\label{eq:hdp_sample_z}
\begin{array}{ll}
&q(z_{di} =  k| \bm{Z}_{t}^{\neg di}, \bm{\lambda_t}, \barpiv)   \propto  \frac{(\alpha \pi_k+C_{dk}^{\neg di}) (C_{kx_{di}}^{\neg di}+\Delta_{kx_{di}}^*) }{\sum_{w}{(C_{kw}^{\neg di}+\Delta_{kw}^*)}} \\
& \exp\!\Big(\!\frac{c y_d(c\epsilon + \lambda_d) \mu_k^*}{n_d \lambda_d}\!-\!\frac{c^2 (\mu_{k}^{*2}+\Sigma_{kk}^* + 2 (\mu_k^* \muv^* + \Sigmav_{\cdot, k}^* )^\top \bm{C}_{d}^{\neg di} ) }{2 n_d^2 \lambda_d}\Big).
\end{array}
\end{equation*}
%where $C_{kw}^{\neg di}$ is the number of times word $w$ is associated with topic $k$, if the i-th word $i$ in the d-th document is excluded.
Besides, for $k > K$ and symmetric Dirichlet prior $\etav$, this becomes $q(z_{di} =  k| \bm{Z}_{t}^{\neg di}, \bm{\lambda_t}, \barpiv) \propto \alpha \pi_k / W$, and therefore the total probability of assigning a new topic is
\setlength\arraycolsep{1pt} \begin{equation*}\label{eq:hdp_sample_z2}
q(z_{di} > K | \Zv_{t}^{\neg di}, \lambdav_t, \barpiv) \propto \alpha \left( 1-\sum\limits_{k = 1}^{K}{\pi_k}\right)/W.
\end{equation*}
\textbf{For $\bm{\lambda}_t$: } The conditional distribution $q(\lambda_d | \Zv_t, \Sv_t, \barpiv)$ is the same as (\ref{eq:lda_sample_lambda}).

\textbf{For $\bm{S}_t$: } The conditional distribution of $s_{dk}$ given $\Zv_t, \barpiv, \\ \lambdav_t$ can be derived from the joint distribution (\ref{eq:joint_zs}):
\begin{equation}
\label{eq:hdp_sample_s}
q(s_{dk} | \Zv_{t}, \lambdav_t, \barpiv) \propto {S(n_d \bar{z}_{dk}, s_{dk}) (\alpha \pi_k)^{s_{dk}}}
\end{equation}
\textbf{For $\bm{\bar{\pi}}$:} It can be derived from (\ref{eq:pamedhdp_zlambdas}) that given $(\Zv_t, \Sv_t, \lambdav_t)$, each $\bar{\pi}_k$ follows the beta distribution, $\bar{\pi}_k \sim \B(a_k, b_k)$, where $a_k = u_k^* + \sum_{d \in B_t} s_{dk}$ and $b_k = v_k^* + \sum_{d \in B_t} \sum_{j > k} s_{dj}$.
%\begin{equation*} %\label{eq:hdp_sample_pi}
%\begin{array}{l}
%q(\bar{\pi}_k | \bm{Z}_t, \bm{S}_t, \lambdav_t)  \propto \bar{\pi}_k^{u_k+\sum_{d \in B_t}{s_{dk}}} (1-\bar{\pi}_k)^{v_k+\sum_{d \in B_t}\sum_{j > k}s_{dj}}
%\end{array}
%\end{equation*}
%Based on $\barpiv$, we then compute $\piv$ via $\pi_k = \bar{\pi}_k \prod_{i < k}(1-\bar{\pi}_{i})$ for all $k \geq 1$.

\iffalse
\begin{algorithm}[t]
\caption{\small{Online MedHDP}}
\label{algo:pamedhdp}
\begin{algorithmic}[1]
\STATE Let $q_0(\bm{\wv}) = \mathcal{N}(0; v^2 I), q_0(\bm{\phi}_k) = \text{Dir}(\gamma)$, $q_0(\bar{\pi}_k) = \text{Beta}(1,\gamma)$, for all $k \geq 0$.
\FOR{$t = 0 \to \infty$}
\STATE Set $q(\wv, \barpiv, \Phiv) = q_t(\wv, \barpiv, \Phiv)$.
\FOR{$i = 1 \to \mathcal{I}$}
\STATE Draw samples $\{\bm{Z}_t^{(j)}, \bm{S}_t^{(j)}, \bm{\lambda}_t^{(j)}\}_{j=1}^{\mathcal{J}}$ with (\ref{eq:hdp_sample_z}, \ref{eq:hdp_sample_z2}, \ref{eq:lda_sample_lambda}, \ref{eq:hdp_sample_pi}, \ref{eq:hdp_sample_s}).
\STATE Update $q(\wv, \barpiv, \Phiv)$ using (\ref{eq:lda_phi2}, \ref{eq:lda_weight2}, \ref{eq:hdp_beta}).
\ENDFOR
\STATE Set $q_{t+1}(\wv, \barpiv, \Phiv) = q(\wv, \barpiv, \Phiv)$.
\ENDFOR
\end{algorithmic}
\end{algorithm}
\fi

Similar to online MedLDA, we iterate the above steps till convergence for training. %The overall paradigm is outlined in Algorithm \ref{algo:pamedhdp}.

\section{Experiments}\label{sec:exp}

We demonstrate the efficiency and prediction accuracy of online MedLDA and MedHDP, denoted as \emph{paMedLDA} and \emph{paMedHDP}, on the 20Newsgroup (20NG) and a large Wikipedia dataset. A sensitivity analysis of the key parameters is also provided. Following the same setting in \cite{zhu2012medlda}, we remove a standard list of stop words. All of the experiments are done on a normal computer with single-core clock rate up to 2.4 GHz.

\subsection{Classification on 20Newsgroup}
\label{sec:mc}

We perform multi-class classification on the entire 20NG dataset with all the 20 categories. The training set contains 11,269 documents, with the smallest category having 376 documents and the biggest category having 599 documents. The test set contains 7,505 documents, with the smallest and biggest categories having 259 and 399 documents respectively. We adopt the "one-vs-all" strategy \cite{rifkin2004defense} to combine binary classifiers for multi-class prediction tasks.

\begin{figure}
\includegraphics[width = 0.5\textwidth]{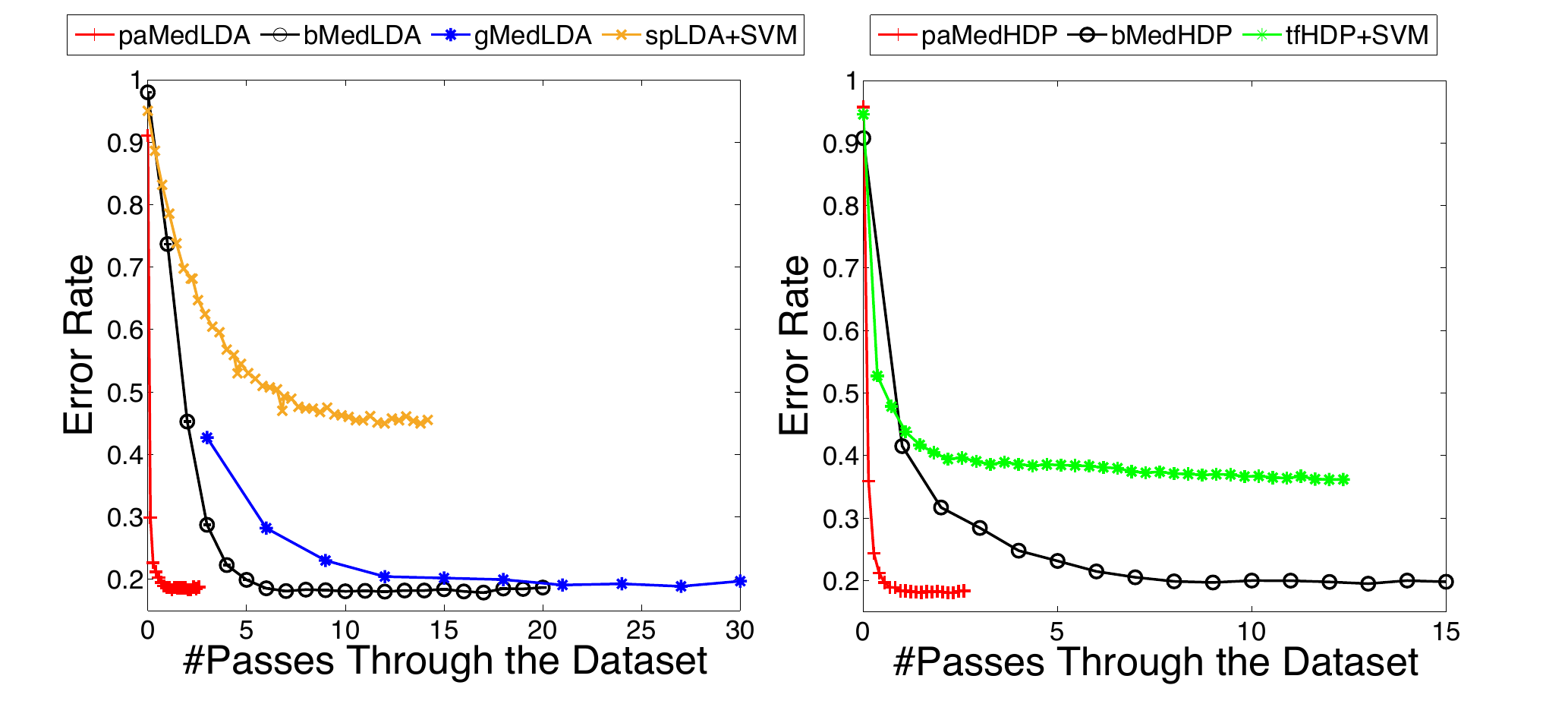}\vspace*{-0.2cm}
\caption{Test errors with different number of passes through the 20NG training dataset. \textbf{Left:} LDA-based models. \textbf{Right:} HDP-based models.}
\vspace*{-0.3cm}
\label{fg:multic_commit}
\end{figure}

We compare paMedLDA and paMedHDP with their batch counterparts, denoted as \emph{bMedLDA} and \emph{bMedHD}P, which are obtained by letting the batch size $|B|$ be equal to the dataset size $D$, and Gibbs MedLDA, denoted as \emph{gMedLDA}, \cite{zhugibbs2013}, which performs Gibbs sampling in the batch manner. We also consider online unsupervised topic models as baselines, including sparse inference for LDA (\emph{spLDA}) \cite{mimno2012sparse}, which has been demonstrated to be superior than online variational LDA \cite{hoffman2010online} in performance, and truncation-free online variational HDP (\emph{tfHDP})~\cite{wang2012truncation}, which has been shown to be promising in nonparametric topic modeling. For both of them, we learn a linear SVM with the topic representations using LIBSVM \cite{chang2011libsvm}. The performances of other batch supervised topic models, such as sLDA \cite{blei2010supervised} and DiscLDA \cite{lacoste2008disclda}, are reported in \cite{zhu2012medlda}.  For all LDA-based topic models, we use symmetric Dirichlet priors $\bm{\alpha} = 1/K \cdot \bm{1}, \bm{\gamma} = 0.5 \cdot \bm{1}$; for all HDP-based topic models, we use $\alpha = 5, \gamma = 1, \bm{\eta} = 0.45 \cdot \bm{1}$; for all MED topic models, we use $\epsilon = 164, c = 1, v = 1$, the choice of which is not crucial to the models' performance as shown in \cite{zhugibbs2013}.

We first analyze how many processed documents are sufficient for each model to converge. Figure \ref{fg:multic_commit} shows the prediction accuracy with the number of passes through the entire 20NG dataset, where $K = 80$ for parametric models and $(\mathcal{I}, \mathcal{J}, \beta) = (1, 2, 0)$ for BayesPA. As we could observe, by solving a series of latent BayesPA learning problems, paMedLDA and paMedHDP fully explore the redundancy of documents and converge in  one pass, while their batch counterparts need many passes as burn-in steps. Besides, compared with the online unsupervised learning algorithms, BayesPA topic models utilize supervising-side information from each mini-batch, and therefore exhibit a faster convergence rate in discrimination ability.
% By solving a series of latent Passive-Aggressive learning problems with hybrid variational inference and data augmentation trick, paMedLDA converges to its batch alternatives in a single pass.
%Besides, among the batch algorithms, bMedLDA  requires fewer passes than gMedLDA, probably because it uses multiple local samples for each global update, thus being more stable.

\begin{figure}[t]
\includegraphics[width = 0.48\textwidth]{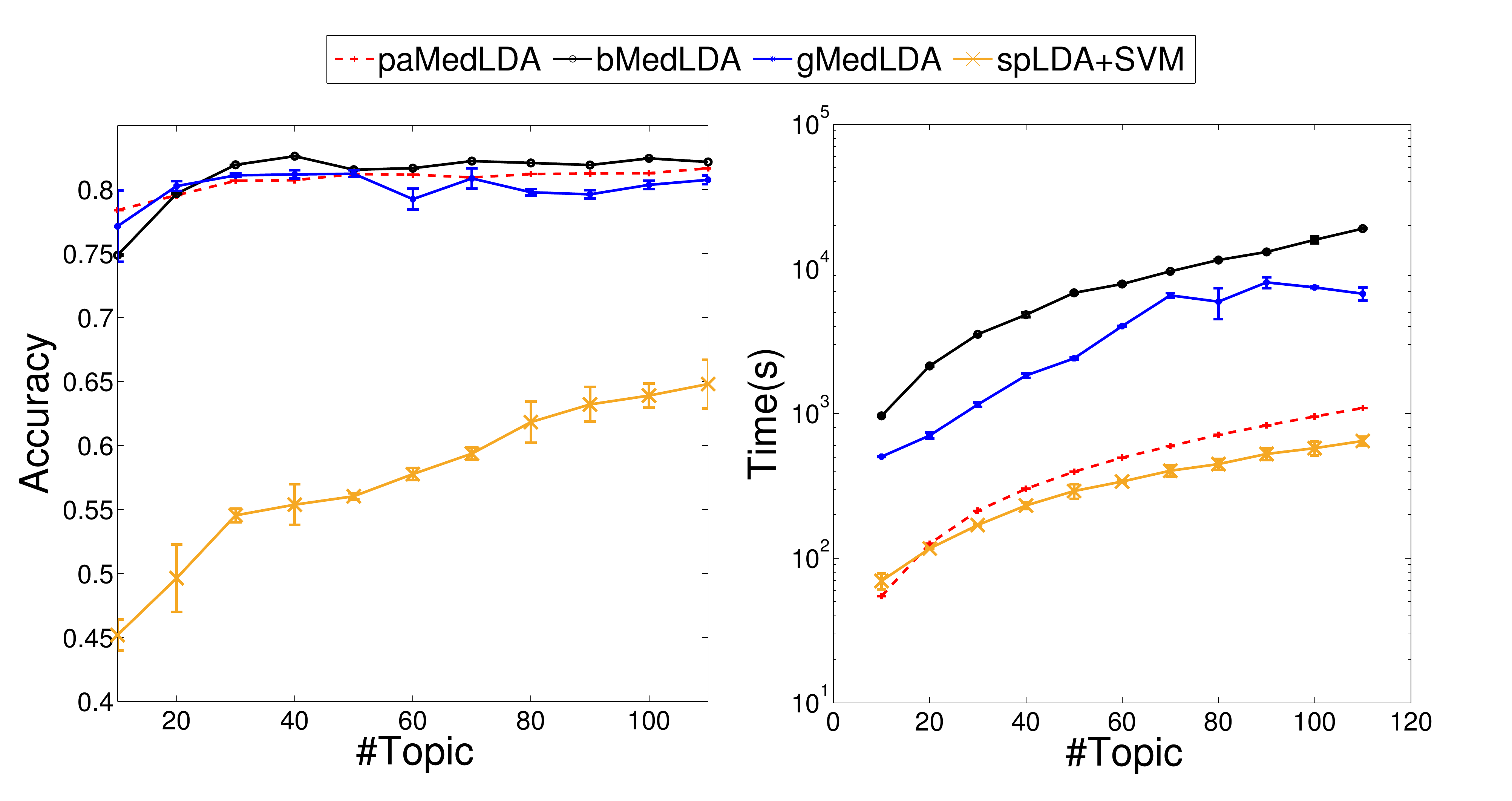}\vspace*{-0.2cm}
\caption{Classification accuracy and running time of paMedLDA and comparison models on the 20NG dataset.}\vspace*{-0.3cm}
\label{fg:multic_topic_lda}
\end{figure}

\begin{figure}[t]
\includegraphics[width = 0.48\textwidth]{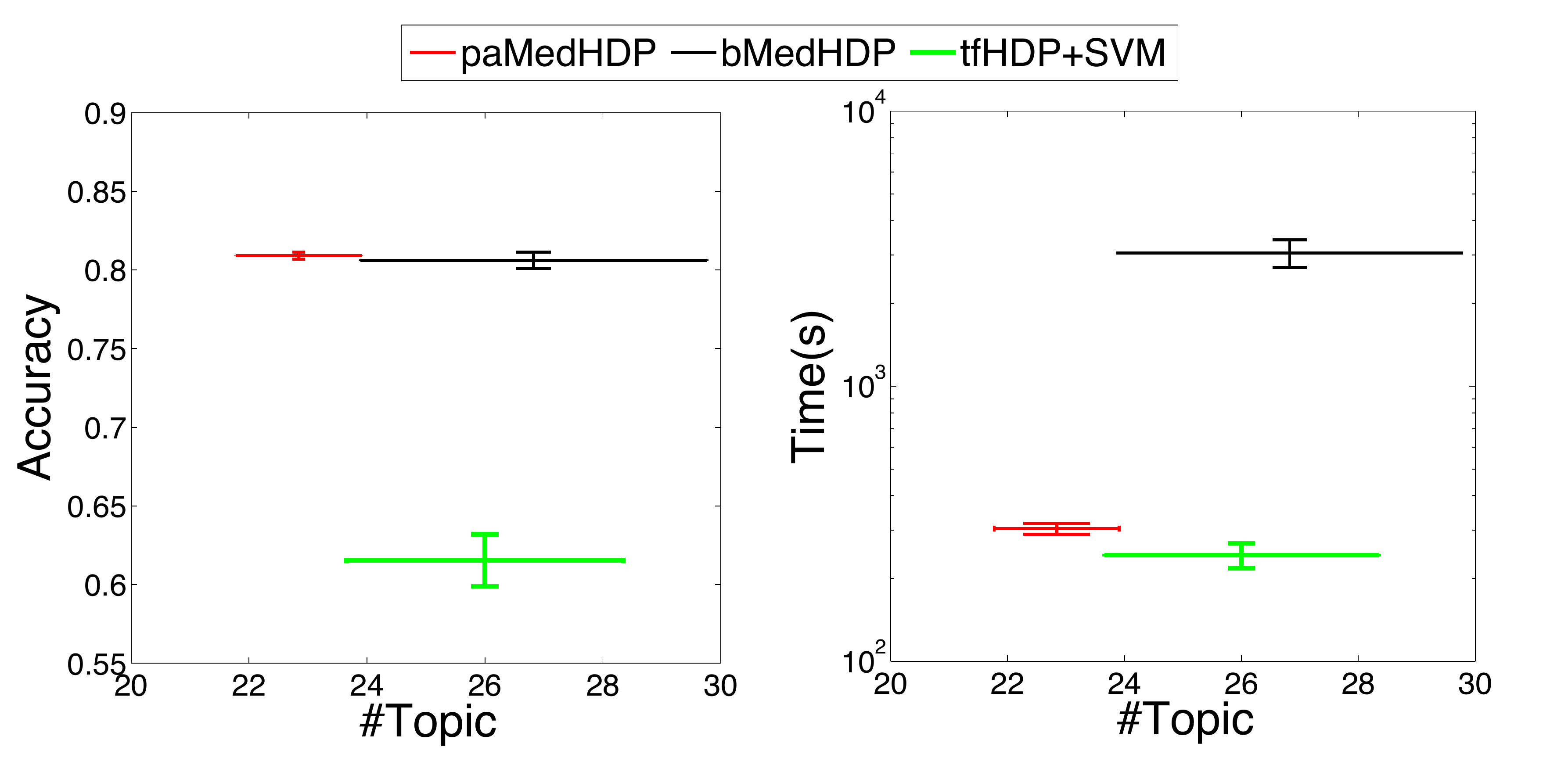}\vspace*{-.2cm}
\caption{Classification accuracy and running time of paMedHDP and comparison models on the 20NG dataset.}\vspace*{-0.2cm}
\label{fg:multic_topic_hdp}
\end{figure}

Next, we study each model's best performance possible and the corresponding training time.
%For test time, paMedLDA adopts the same testing approach as Gibbs MedLDA,  \cite{zhugibbs2013} provides a detailed comparison of test time.
To allow for a fair comparison, we train each model until the relative change of  its objective is less than $10^{-4}$.
%$\mathcal{M} = 8$ for bMedLDA, $\mathcal{M} = 30$ for gMedLDA, and number of passes $\mathcal{E} = 3$ for oLDA, $\mathcal{E} = 10$ for tfHDP.
%In particular, we use a single pass for paMedLDA.
 Figure \ref{fg:multic_topic_lda} shows the accuracy and training time of LDA-based models on the whole dataset with varying numbers of topics. Similarly, Figure \ref{fg:multic_topic_hdp} shows the accuracy and training time of HDP-based models, where the dots stand for the mean inferred numbers of topics, and the lengths of the horizontal bars represent their standard deviations. As we can see, BayesPA topic models, at the power of online learning, are about 1 order of magnitude faster than their batch counterparts in training time. Furthermore, thanks to the merits of Gibbs sampling, which does not pose strict mean-field assumptions about the independence of latent variables, BayesPA topic models parallel their batch alternatives in accuracy.

% The result is produced using burn-in steps $\mathcal{M} = 10$ for bMedHDP, number of passes $\mathcal{E} = 10$ for tfHDP and a single pass for paMedHDP.

%Comparing Figure \ref{fg:multic_topic_lda} and Figure \ref{fg:multic_topic_hdp}, we could see the test accuracies of paMedLDA and paMedHDP are close, while paMedHDP requires less time to train given the same number of topics. A possible explanation is that paMedHDP handles fewer topics at the first several iterations, and hence becomes faster.

%One more thing to notice is the preferable number of topics. As shown in Figure \ref{fg:multic_topic_lda}, the performance of MedLDA varies little for a large interval of \#topics (for example $K \in [20, 110]$). Therefore, any topic number $K$ in that interval is preferable by an MedHDP model. From Figure \ref{fg:multic_topic_hdp}, we see that paMedHDP infers around 23 topics, while the batch counterparts

\subsection{Further Discussions}

We provide further discussions on BayesPA learning for topic models. First, we analyze the models' sensitivity to some  key parameters. Second, we illustrate an application with a large Wikipedia dataset containing 1.1 million documents, where class labels are not exclusive.
\label{sec:discuss}

\subsubsection{Sensitivity Analysis}

\begin{figure}[t]
\includegraphics[width = 0.48\textwidth]{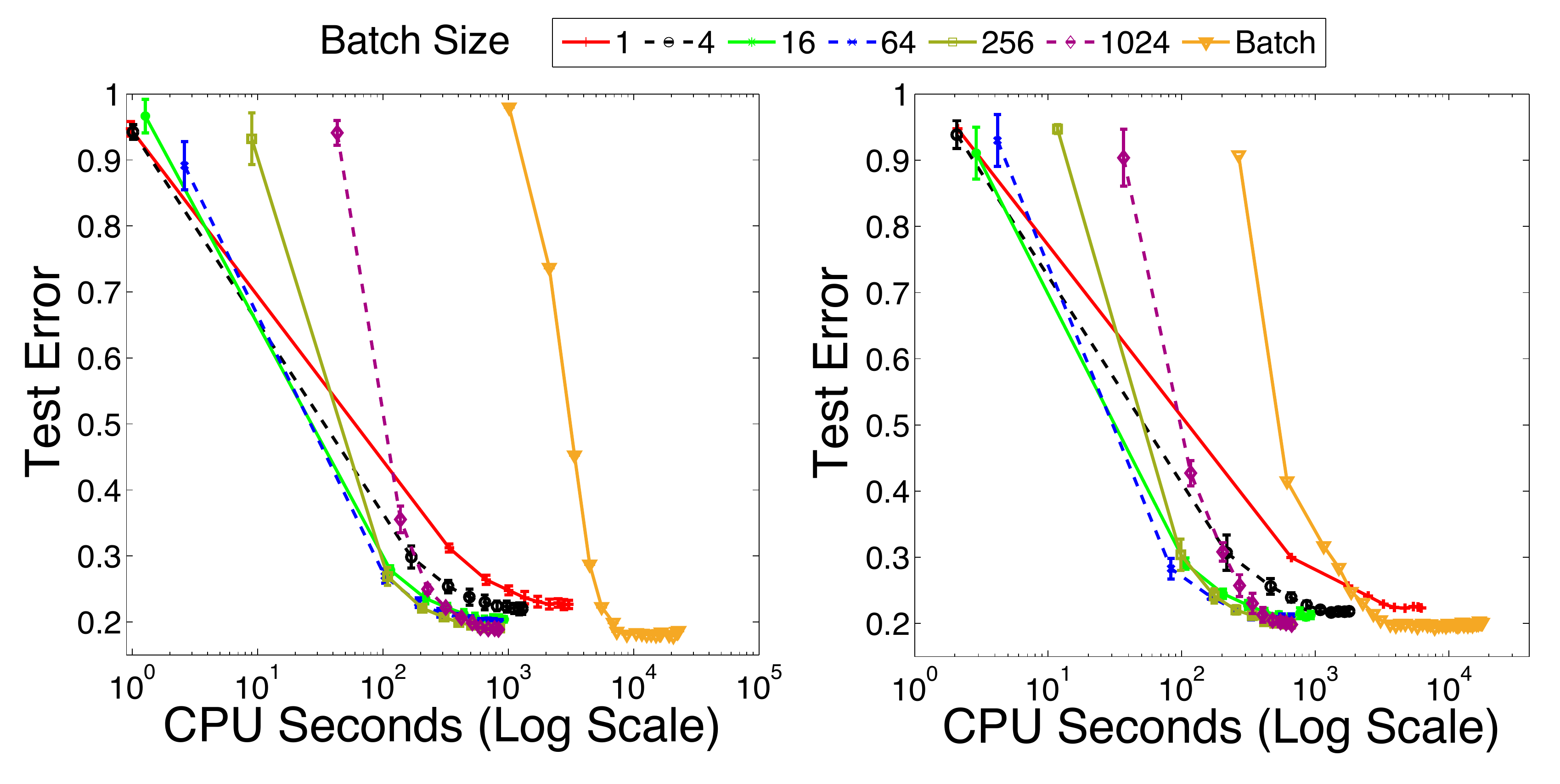}\vspace*{-0.2cm}
\caption{Test errors of paMedLDA (left) and paMedHDP (right) with different batch sizes on the 20NG dataset.}\vspace*{-0.3cm}
\label{fg:multic_batchsize}
\end{figure}

\textbf{Batch Size $|B|$: } Figure \ref{fg:multic_batchsize} presents the test errors of BayesPA topic models as a function of training time on the entire 20NG dataset with various batch sizes, where $K = 40$. We can see that the convergence speeds of different algorithms vary. First of all, the batch algorithms suffer from multiple passes through the dataset and therefore are much slower than the online alternatives. Second, we could observe that algorithms with medium batch sizes ($|B| = 64, 256$) converge faster. If we choose a batch size too small, for example, $|B| = 1$, each iteration would not provide sufficient evidence for the update of global variables; if the batch size is too large, each mini-batch becomes redundant and the convergence rate reduces.

\textbf{Number of iterations $\mathcal{I}$ and samples $\mathcal{J}$: } Since the time complexity of Algorithm \ref{algo:pamedlda} is linear in both $\mathcal{I}$ and $\mathcal{J}$, we would like to know how these parameters influence the quality of the trained model.  First, notice that the first $\beta$ samples are discarded as burn-in steps. To understand how large $\beta$ is sufficient, we consider the settings of the pairs $(\mathcal{J}, \beta)$ and check the prediction accuracy of Algorithm \ref{algo:pamedlda} for $K = 40, |B| = 512$, as shown in Table \ref{tb:samplen}.

% note: The background color of cell is obtained by renormalizing the accuracy into [0.4, 0.8]
\begin{table}[h]
\caption{Effect of the number of samples and burn-in steps.}
\label{tb:samplen}
\begin{center}
\begin{tabular}{c  c  c  c c c }
\backslashbox[5mm]{$\mathcal{J}$}{$\beta$} & 0 & 2 & 4 & 6 & 8 \\
\hline
1 & \cellcolor[gray]{0.4} 0.783 &\\
3 & \cellcolor[gray]{0.72} 0.803 & \cellcolor[gray]{0.6560} 0.799 \\
5  &\cellcolor[gray]{0.8} 0.808 &\cellcolor[gray]{0.72} 0.803 & \cellcolor[gray]{0.5440} 0.792 \\
9 & \cellcolor[gray]{0.761} 0.806 & \cellcolor[gray]{0.761} 0.806 & \cellcolor[gray]{0.761} 0.806 & \cellcolor[gray]{0.730} 0.804 & \cellcolor[gray]{0.608} 0.796\\
\hline
\end{tabular}
\end{center}
\end{table}

We can see that accuracies closer to the diagonal of the table are relatively lower, while settings with the same number of kept samples, e.g. $(\mathcal{J}, \beta) = (3,0), (5,2), (9,6)$, yield similar results. The number of kept samples exhibits a more significant role in the performance of BayesPA topic models than the burn-in steps.

\begin{figure}[t]
\includegraphics[width = 0.5\textwidth]{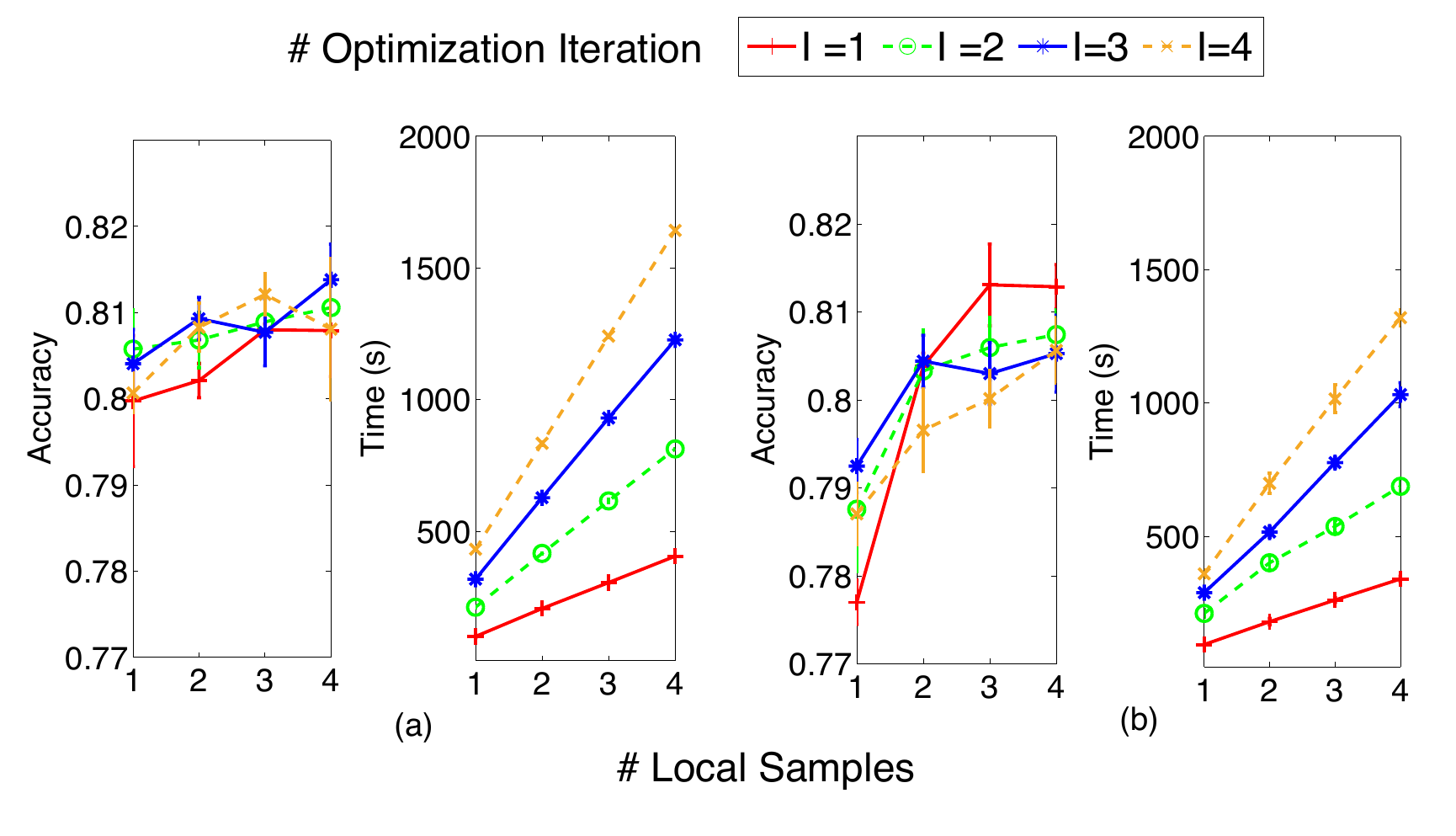}
\vspace*{-0.5cm}
\caption{Classification accuracies and training time of \textbf{(a)}: paMedLDA, \textbf{(b):} paMedHDP, with different combinations of $(\mathcal{I}, \mathcal{J})$ on the 20NG dataset.}
\label{fg:multic_IJ}
\end{figure}

Next, we analyze which setting of $(\mathcal{I}, \mathcal{J})$ guarantees good performance. Figure \ref{fg:multic_IJ} presents the results. As we can see, for $\mathcal{J} = 1$, the algorithms suffer from the noisy approximation and therefore sacrifices prediction accuracy. But for larger $\mathcal{J}$, simply $\mathcal{I} = 1$ is promising, possibly due to the redundancy among mini-batches.

\subsubsection{Multi-Task Classification}

For multi-task classification, a set of binary classifiers are trained, each of which identifies whether a document $\bm{x}_d$ belongs to a specific task/category $\bm{y}_d^{\tau} \in \{+1, -1\}$. These binary classifiers are allowed to share common latent representations and therefore could be attained via a modified BayesPA update equation:
\begin{equation*}
\label{eq:onlinepa_latent_mt}
\underset{q \in \mathcal{F}_t}{\operatorname{min}}{~\mathcal{L}(q(\bm{w}, \bm{\mathcal{M}}, \Hv_t))+ 2 c \sum\limits_{\tau = 1}^{\mathcal{T}} \ell_\epsilon(q(\bm{w}, \bm{\mathcal{M}}, \Hv_t); \bm{X}_t, \bm{Y}_t^\tau)}
\end{equation*}
where $\mathcal{T}$ is the total number of tasks. We can then derive the multi-task version of Passive-Aggressive topic models, denoted by \emph{paMedLDA-mt} and \emph{paMedHDP-mt}, in a way similar to Section \ref{sec:pamedlda} and \ref{sec:pamedhdp}.

We test paMedLDA-mt and paMedHDP-mt as well as comparison models, including bMedLDA-mt, bMedHDP-mt and gMedLDA-mt \cite{zhu2013scalable} on a large Wiki dataset built from the Wikipedia set used in PASCAL LSHC challenge 2012 \footnote{See http://lshtc.iit.demokritos.gr/.}. The Wiki dataset is a collection of documents with labels up to 20 different kinds, while the data distribution among the labels is balanced.  The training/testing split is 1.1 million / 5 thousand. To measure performance, we use F1 score, the harmonic mean of precision and recall.

\begin{figure}
\includegraphics[width = 0.48\textwidth]{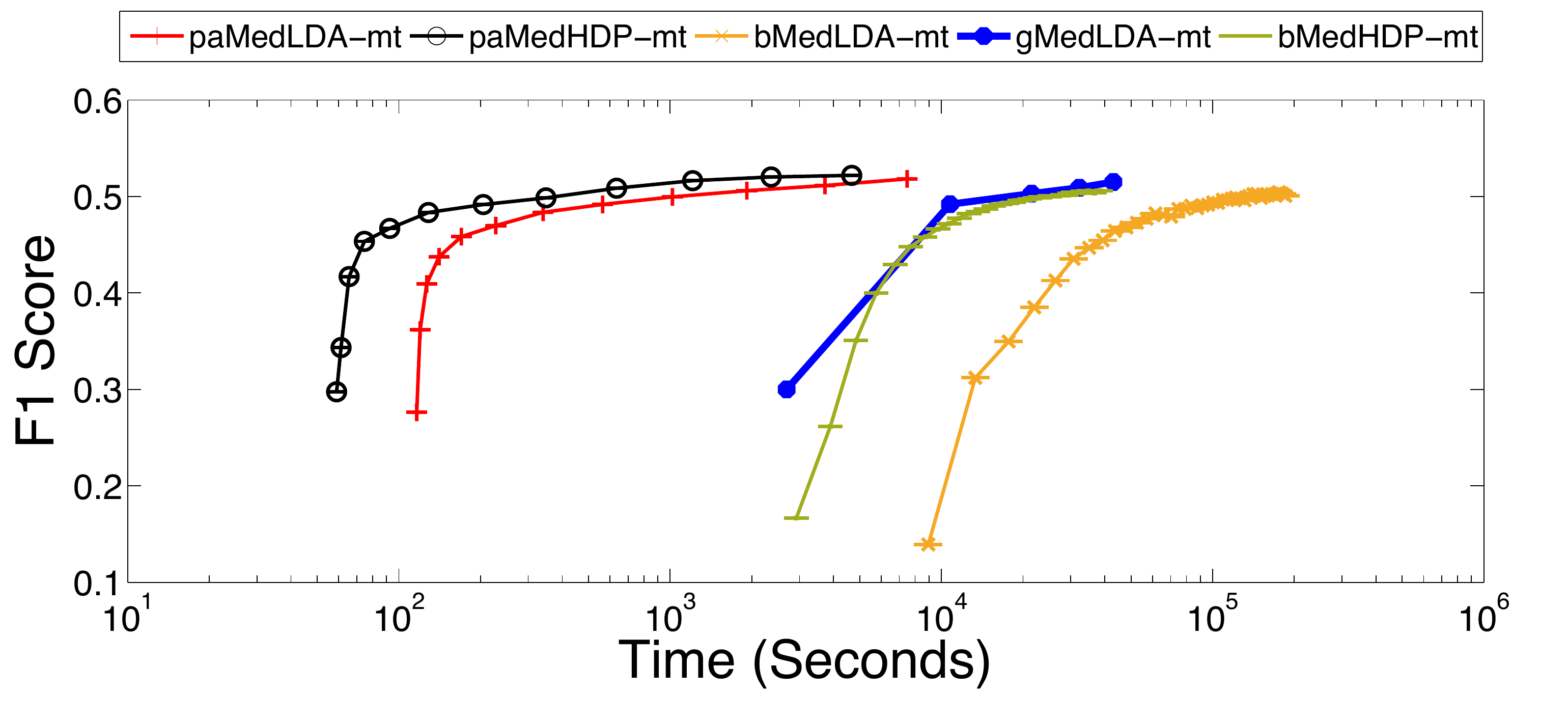}
\vspace*{-0.5cm}
\caption{F1 scores of various models on the 1.1M wikipedia dataset.}
\vspace*{-0.5cm}
\label{fg:multic_mtask_commit}
\end{figure}

Figure \ref{fg:multic_mtask_commit} shows the F1 scores of various models as a function of training time. We can see that BayesPA topic models are again about 1 order of magnitude faster than the batch alternatives and yet produce comparable results. Therefore, BayesPA topic models are potentially extendable to large-scale multi-class settings.

%In fact, the similarity of multi-task Bayesian PA topic models and multi-task Gibbs MedLDA suggests that a highly paralleled and scalable implementation is possible \cite{zhu2013scalable}.

\section{Conclusions and Future Work}
\label{sec:con}

We present online Bayesian Passive-Aggressive (BayesPA)  learning as a new framework for max-margin Bayesian inference of online streaming data. We show that BayesPA subsumes the online PA, and more significantly, generalizes naturally to incorporate latent variables and to perform nonparametric Bayesian inference, therefore providing great flexibility for explorative analysis. Based on the ideas of BayesPA, we develop efficient online learning algorithms for max-margin topic models as well as their nonparametric extensions. Empirical experiments on several real datasets demonstrate significant improvements on time efficiency, while maintaining comparable results.

\iffalse
Conceptually, we generalize the online Passive-Aggressive algorithms by \cite{crammer2006pa}, with new techniques developed to deal with latent variables. Empirically, we apply Bayesian PA learning to maximum entropy discriminant topic models and propose effective variational inference methods. Experiments on several real datasets demonstrate our approach yield results comparable to batch counterparts, while being significantly faster.
\fi

As future work, we are interested in showing provable bounds in its convergence and limitations. Furthermore, better understanding its mathematical structure would allow one to design more involved BayesPA algorithms for various models. We are also interested in developing highly scalable, distributed \cite{broderick2013streaming} BayesPA learning paradigms, which will better meet the demand of processing massive real data available today.
% In the unusual situation where you want a paper to appear in the
% references without citing it in the main text, use \nocite
%\nocite{langley00}

\bibliography{bpal}

\begin{thebibliography}{33}
\providecommand{\natexlab}[1]{#1}
\providecommand{\url}[1]{\texttt{#1}}
\expandafter\ifx\csname urlstyle\endcsname\relax
  \providecommand{\doi}[1]{doi: #1}\else
  \providecommand{\doi}{doi: \begingroup \urlstyle{rm}\Url}\fi

\bibitem[Ahn et~al.(2012)Ahn, Korattikara, and Welling]{welling2012mc}
Ahn, S., Korattikara, A., and Welling, M.
\newblock {Bayesian posterior sampling via stochastic gradient Fisher scoring}.
\newblock In \emph{ICML}, 2012.

\bibitem[Antoniak(1974)]{antoniak1974mixtures}
Antoniak, C.E.
\newblock {Mixtures of Dirichlet processes with applications to Bayesian
  nonparametric problems}.
\newblock \emph{The annals of statistics}, pp.\  1152--1174, 1974.

\bibitem[Blei \& McAuliffe(2010)Blei and McAuliffe]{blei2010supervised}
Blei, D.M. and McAuliffe, J.D.
\newblock {Supervised topic models}.
\newblock In \emph{NIPS}, 2010.

\bibitem[Blei et~al.(2003)Blei, Ng, and Jordan]{blei2003lda}
Blei, D.M., Ng, A., and Jordan, M.I.
\newblock Latent {Dirichlet} allocation.
\newblock \emph{JMLR}, 3:\penalty0 993--1022, 2003.

\bibitem[Broderick et~al.(2013)Broderick, Boyd, Wibisono, Wilson, and
  Jordan]{broderick2013streaming}
Broderick, T., Boyd, N., Wibisono, A., Wilson, A.C., and Jordan, M.I.
\newblock {Streaming variational Bayes}.
\newblock \emph{arXiv preprint arXiv:1307.6769}, 2013.

\bibitem[Chang \& Lin(2011)Chang and Lin]{chang2011libsvm}
Chang, C.C. and Lin, C.-J.
\newblock {LIBSVM: a library for support vector machines}.
\newblock \emph{ACM Transactions on Intelligent Systems and Technology (TIST)},
  2\penalty0 (3):\penalty0 27, 2011.

\bibitem[Chiang et~al.(2008)Chiang, Marton, and Resnik]{chiang2008emnlp}
Chiang, D., Marton, Y., and Resnik, P.
\newblock {Online large-margin training of syntactic and structural translation
  features}.
\newblock In \emph{EMNLP}, 2008.

\bibitem[Cortes \& Vapnik(1995)Cortes and Vapnik]{cortes1995support}
Cortes, C. and Vapnik, V.
\newblock Support-vector networks.
\newblock \emph{Machine learning}, 20\penalty0 (3):\penalty0 273--297, 1995.

\bibitem[Crammer et~al.(2006)Crammer, Dekel, Keshet, Shalel-Shwartz, and
  Singer]{crammer2006pa}
Crammer, K., Dekel, O., Keshet, J., Shalel-Shwartz, S., and Singer, Y.
\newblock {Online Passive-Aggressive learning}.
\newblock \emph{JMLR}, 7:\penalty0 551--585, 2006.

\bibitem[Devroye(1986)]{devroye1986sample}
Devroye, L.
\newblock \emph{Non-Uniform Random Variate Generation}.
\newblock Springer, 1986.

\bibitem[Ghahramani \& Griffiths(2005)Ghahramani and Griffiths]{Griffiths:tr05}
Ghahramani, Z. and Griffiths, T.L.
\newblock {Infinite latent feature models and the Indian buffet process}.
\newblock In \emph{NIPS}, 2005.

\bibitem[Griffiths \& Steyvers(2004)Griffiths and
  Steyvers]{griffiths2004finding}
Griffiths, T.L. and Steyvers, M.
\newblock Finding scientific topics.
\newblock \emph{PNAS}, 101:\penalty0 5228--5235, 2004.

\bibitem[Hjort(2010)]{hjort2010bayesian}
Hjort, N.L.
\newblock \emph{{Bayesian Nonparametrics}}.
\newblock Cambridge University Press, 2010.

\bibitem[Hoffman et~al.(2010)Hoffman, Bach, and Blei]{hoffman2010online}
Hoffman, M., Bach, F.R., and Blei, D.M.
\newblock Online learning for latent {Dirichlet} allocation.
\newblock In \emph{NIPS}, 2010.

\bibitem[Jaakkola et~al.(1999)Jaakkola, Meila, and Jebara]{Jaakkola:99}
Jaakkola, T., Meila, M., and Jebara, T.
\newblock Maximum entropy discrimination.
\newblock In \emph{NIPS}, 1999.

\bibitem[Jebara(2011)]{jebara2011mtl}
Jebara, T.
\newblock {Multitask sparsity via maximum entropy discrimination}.
\newblock \emph{JMLR}, 12:\penalty0 75--110, 2011.

\bibitem[Jiang et~al.(2012)Jiang, Zhu, Sun, and Xing]{jiang2012monte}
Jiang, Q., Zhu, J., Sun, M., and Xing, E.P.
\newblock {Monte Carlo Methods for Maximum Margin Supervised Topic Models}.
\newblock In \emph{NIPS}, 2012.

\bibitem[Jordan et~al.(1998)Jordan, Ghahramani, Jaakkola, and
  Saul]{jordan1998introduction}
Jordan, M.I., Ghahramani, Z., Jaakkola, T.S., and Saul, L.K.
\newblock \emph{An Introduction to Variational Methods for Graphical Models}.
\newblock Springer, 1998.

\bibitem[Lacoste-Julien et~al.(2008)Lacoste-Julien, Sha, and
  Jordan]{lacoste2008disclda}
Lacoste-Julien, S., Sha, F., and Jordan, M.I.
\newblock {DiscLDA: Discriminative learning for dimensionality reduction and
  classification}.
\newblock In \emph{NIPS}, 2008.

\bibitem[McDonald et~al.(2005)McDonald, Crammer, and
  Pereira]{McDonald2005parsing}
McDonald, R., Crammer, K., and Pereira, F.
\newblock Online large-margin training of dependency parsers.
\newblock In \emph{ACL}, 2005.

\bibitem[Michael et~al.(1976)Michael, Schucany, and Haas]{Michael:IG76}
Michael, J.R., Schucany, W.R., and Haas, R.W.
\newblock Generating random variates using transformations with multiple roots.
\newblock \emph{The American Statistician}, 30\penalty0 (2):\penalty0 88--90,
  1976.

\bibitem[Mimno et~al.(2012)Mimno, Hoffman, and Blei]{mimno2012sparse}
Mimno, D., Hoffman, M., and Blei, D.M.
\newblock Sparse stochastic inference for latent dirichlet allocation.
\newblock \emph{ICML}, 2012.

\bibitem[Rifkin \& Klautau(2004)Rifkin and Klautau]{rifkin2004defense}
Rifkin, R. and Klautau, A.
\newblock In defense of one-vs-all classification.
\newblock \emph{JMLR}, 5:\penalty0 101--141, 2004.

\bibitem[Teh et~al.(2006{\natexlab{a}})Teh, Jordan, Beal, and
  Blei]{teh2006hierarchical}
Teh, Y.W., Jordan, M.I., Beal, M.J., and Blei, D.M.
\newblock {Hierarchical Dirichlet processes}.
\newblock \emph{Journal of the American Statistical Association}, 101\penalty0
  (476), 2006{\natexlab{a}}.

\bibitem[Teh et~al.(2006{\natexlab{b}})Teh, Newman, and
  Welling]{teh2006collapsed}
Teh, Y.W., Newman, D., and Welling, M.
\newblock {A collapsed variational Bayesian inference algorithm for latent
  Dirichlet allocation}.
\newblock In \emph{NIPS}, 2006{\natexlab{b}}.

\bibitem[Wang \& Blei(2012)Wang and Blei]{wang2012truncation}
Wang, C. and Blei, D.M.
\newblock {Truncation-free online variational inference for Bayesian
  nonparametric models}.
\newblock In \emph{NIPS}, 2012.

\bibitem[Wang et~al.(2011)Wang, Paisley, and Blei]{wang2011onlinealgo}
Wang, C., Paisley, J.W., and Blei, D.M.
\newblock {Online variational inference for the hierarchical Dirichlet
  process}.
\newblock In \emph{AISTATS}, 2011.

\bibitem[Xu et~al.(2012)Xu, Zhu, and Zhang]{Xu:NIPS12}
Xu, M., Zhu, J., and Zhang, B.
\newblock Bayesian nonparametric maximum margin matrix factorization for
  collaborative prediction.
\newblock In \emph{NIPS}, 2012.

\bibitem[Zhu \& Xing(2009)Zhu and Xing]{Zhu:jmlr09}
Zhu, J. and Xing, E.P.
\newblock Maximum entropy discrimination {Markov} networks.
\newblock \emph{JMLR}, 10:\penalty0 2531--2569, 2009.

\bibitem[Zhu et~al.(2011)Zhu, Chen, and Xing]{zhu2011infinite}
Zhu, J., Chen, N., and Xing, E.P.
\newblock {Infinite SVM: a Dirichlet process mixture of large-margin kernel
  machines}.
\newblock In \emph{ICML}, 2011.

\bibitem[Zhu et~al.(2012)Zhu, Ahmed, and Xing]{zhu2012medlda}
Zhu, J., Ahmed, A., and Xing, E.P.
\newblock {MedLDA: maximum margin supervised topic models}.
\newblock \emph{JMLR}, 13:\penalty0 2237--2278, 2012.

\bibitem[Zhu et~al.(2013{\natexlab{a}})Zhu, Chen, Perkins, and
  Zhang]{zhugibbs2013}
Zhu, J., Chen, N., Perkins, H., and Zhang, B.
\newblock Gibbs max-margin topic models with fast sampling algorithms.
\newblock \emph{ICML}, 2013{\natexlab{a}}.

\bibitem[Zhu et~al.(2013{\natexlab{b}})Zhu, Zheng, Zhou, and
  Zhang]{zhu2013scalable}
Zhu, J., Zheng, X., Zhou, L., and Zhang, B.
\newblock Scalable inference in max-margin topic models.
\newblock In \emph{SIGKDD}, 2013{\natexlab{b}}.

\end{thebibliography}
\bibliographystyle{icml2014}

\section*{Appendix A: Proof of Lemma \ref{lm:subsume}}\label{app:subsume}

In this section, we prove Lemma \ref{lm:subsume}. We should note that our deviations below also provide insights for the developments of online BayesPA algorithms with the averaging classifiers.
\begin{proof}
 We prove for the more generalized soft-margin version of BayesPA learning, which can be reformulated using a slack variable $\xi$:
\setlength\arraycolsep{1pt} \begin{equation}\label{eq:onlinepa_reg_app}
\begin{array}{ccc}
& q_{t+1}(\bm{w}) = & \underset{q(\bm{w}) \in \mathcal{P}}{\operatorname{argmin}} ~\text{KL}[q(\wv) || q_t(\wv)] +c \xi \\
&& \text{s.t.}: y_t \mathbb{E}_q[{\wv}^\top \xv_t] \geq \epsilon-\xi, ~~ \xi \geq 0.
\end{array}
\end{equation}
Similar to Corollary 5 in \cite{zhu2012medlda}, the optimal solution $q^*(\wv)$ of the above problem can be derived from its functional Lagrangian  and has the following form:
\begin{equation}
\label{eq:optimal}
q^*(\wv) = \frac{1}{\Gamma(\tau)} q_t(\wv) \exp(\tau 	y_t \wv^\top \xv_t)
\end{equation}
where $\Gamma(\tau)$ is a normalization term and $\tau$ is the optimal solution to the dual problem:
\begin{equation}
\begin{array}{rl}
\label{eq:dual}
\max\limits_{\tau} & {~\tau \epsilon-\log \Gamma(\tau)} \\
\text{s.t. } &  0 \leq \tau \leq c
\end{array}
\end{equation} 
Using this primal-dual interpretation, we first prove that for prior $p_0(\bm{w}) = \mathcal{N}(\bm{w}_0, I)$, $q_t(\bm{w}) = \mathcal{N}(\muv_t, I)$ for some $\muv_t$ in each round $t = 0, 1, 2, ...$. This can be shown by induction. Assume for round $t$, the distribution $q_{t}(\wv) = \mathcal{N}(\muv_{t}, I)$. Then for round $t+1$, the distribution by (\ref{eq:optimal}) is
\begin{equation}
q_{t+1}(\wv) = \frac{1}{\mathcal{C} \cdot \Gamma(\tau)} \exp\Big(-\frac{1}{2} ||\wv-(\muv_t+\tau y_t \xv_t)||^2\Big)
\end{equation}
where $\mathcal{C}$ is some constant. Therefore, the distribution $q_{t+1}(\wv) = \mathcal{N}(\mu_t+\tau \xv_t, I)$. As a by-product, the normalization term $\Gamma(\tau) = \sqrt{2 \pi} \exp(\tau y_t \xv_t^\top \muv_t+\frac{1}{2} \tau^2 \xv_t^\top \xv_t)$.

Next, we show that $\muv_{t+1} = \muv_{t}+\tau y_t \xv_t$ is the optimal solution of the online Passive-Aggressive update rule \cite{crammer2006pa}. To see this, we plug the derived $\Gamma(\tau)$ into (\ref{eq:dual}), and obtain
\begin{equation}
\begin{array}{rl}
\label{eq:dualx}
\max\limits_{\tau}& {~\tau-\frac{1}{2}\tau^2 \xv_t^\top \xv_t-\tau y_t \muv_t^\top \xv_t} \\
\text{s.t. }&  0 \leq \tau \leq c
\end{array}
\end{equation}
which is exactly the dual form of the online Passive-Aggressive update rule:
\begin{equation}
\begin{array}{rl}
\muv_{t+1}^* = & \arg\min{~ ||\muv-\muv_{t}||^2+c \xi } \\
\text{s.t.   }  & y_t \muv^\top \xv_t \geq \epsilon-\xi, ~~ \xi \geq 0,
\end{array}
\end{equation}
the optimal solution to which is $\muv_{t+1}^* = \muv_{t}+\tau y_t \xv_t$. It is then clear that  $\mu_{t+1} = \mu_{t+1}^*$.
\end{proof}

\section*{Appendix B: }\label{app:upperbound}
We show the objective in (\ref{eq:onlinepa_augmented}) is an upper bound of that in (\ref{eq:onlinepa_latent}), that is,
%\begin{lemma}
%The following inequality holds true:
\begin{equation} \label{eq:varbound}
\begin{array}{l}
\mathcal{L}(q( \wv, \Phiv, \Zv_t, \lambdav_t)) -\mathbb{E}_q[\log(\psi(\Yv_t, \lambdav_t | \Zv_t, \wv))] \\\\
~~~ \geq \mathcal{L}(q( \wv, \Phiv, \Zv_t))+2 c \sum\limits_{d \in B_t}{\mathbb{E}_q[(\xi_d)_+]}
\end{array}
\end{equation}
where $\mathcal{L}(q) = \KL[q || q_t(\wv, \Phiv) q_0(\Zv_t)]$.
%\end{lemma}

\begin{proof}
We first have
\begin{equation*}
\mathcal{L}(q(\wv, \Phiv, \Zv_t, \lambdav_t)) = \mathbb{E}_q[\log \frac{q(\lambdav_t ~|~ \wv, \Phiv, \Zv_t) q(\wv, \Phiv, \Zv_t)}{q_t(\wv, \Phiv, \Zv_t)}],
\end{equation*}
and
\begin{equation*}
\mathcal{L}(q(\wv, \Phiv, \Zv_t)) = \mathbb{E}_q[\log \frac{q(\wv, \Phiv, \Zv_t)}{q_t(\wv, \Phiv, \Zv_t)}]
\end{equation*}
Comparing these two equations and canceling out common factors, we know that in order for (\ref{eq:varbound}) to make sense, it suffices to prove
\begin{equation}\label{eq:boundlemma_entropy}
\mathbb{H}[q']-\mathbb{E}_{q'}[\log(\psi(\Yv_t, \lambdav_t | \Zv_t, \wv)] \geq 2 c \sum\limits_{d \in B_t}{\mathbb{E}_{q'}[(\xi_d)_+]}
\end{equation}
is uniformly true for any given $(\wv, \Phiv, \Zv_t)$, where $\mathbb{H}(\cdot)$ is the entropy operator and $q' = q(\lambdav_t ~|~ \wv, \Phiv, \Zv_t)$. The inequality (\ref{eq:boundlemma_entropy}) can be reformulated as
\begin{equation} \label{eq:boundlemma}
\mathbb{E}_{q'}[\log \frac{q'}{\psi(\Yv_t, \lambdav_t | \Zv_t, \wv)}] \geq 2c \sum\limits_{d \in B_t}{\mathbb{E}_{q'}[(\xi_d)_+]}
\end{equation}
Exploiting the convexity of the function $\log(\cdot)$, i.e.
\small
\begin{equation*}
-\mathbb{E}_{q'}[\log \frac{\psi(\Yv_t, \lambdav_t | \Zv_t, \wv)}{q'}]  \geq -\log \int_{\lambdav_t}{\psi(\Yv_t, \lambdav_t | \Zv_t, \wv) ~d\lambdav_t},
\end{equation*}
\normalsize
and utilizing the equality (\ref{eq:scalemix}), we then have (\ref{eq:boundlemma}) and therefore prove (\ref{eq:varbound}).
\end{proof}

\end{document}